\documentclass[10pt,twocolumn,letterpaper]{article}

\usepackage[pagenumbers]{iccv} 

\definecolor{iccvblue}{rgb}{0.21,0.49,0.74}
\usepackage[pagebackref,breaklinks,colorlinks,allcolors=iccvblue]{hyperref}

\def\paperID{7172} 
\def\confName{ICCV}
\def\confYear{2025}

\usepackage{hyperref}
\usepackage{algorithm}
\usepackage{algorithmic}
\usepackage{amsmath}
\usepackage{tikz}
\usepackage{color, colortbl}
\usepackage{xcolor}
\usepackage{CJKutf8}
\usepackage{booktabs}
\usepackage{cuted}
\usepackage[accsupp]{axessibility}

\usepackage{multirow}
\usepackage{adjustbox}
\usepackage{makecell}

\definecolor{myblue}{rgb}{0.88,0.98,1}
\definecolor{mygreen}{rgb}{0.92, 1.0, 0.92}
\definecolor{myred}{rgb}{1, 0.9, 0.9}
\definecolor{mygray}{gray}{0.95}

\definecolor{mygreen2}{rgb}{0.30, 0.75, 0.43}
\definecolor{mygray2}{gray}{0.3}

\definecolor{posterblue}{rgb}{0.61,0.89, 0.96}
\definecolor{chartgreen}{rgb}{0.89,0.97, 0.67}
\definecolor{pdfred}{rgb}{1,0.86, 0.88}

\title{
\vspace{-0.5cm}

\begin{minipage}{.04\textwidth}
\centering
\includegraphics[width=0.8\linewidth]{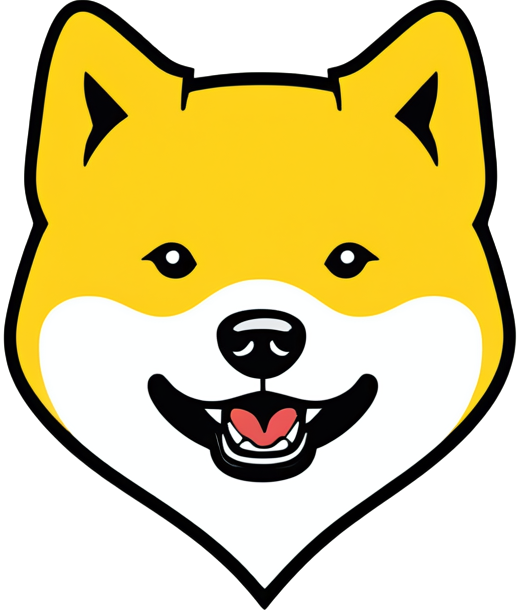} 
\end{minipage}
DOGR: Towards Versatile Visual Document Grounding and Referring
\vspace{-0.7cm}
}

\makeatletter
\def\thanks#1{\protected@xdef\@thanks{\@thanks
        \protect\footnotetext[0]{#1}}}
\makeatother

\author{
  Yinan Zhou$^{1,2,3*}$, 
  Yuxin Chen$^{2*}$$^\dagger$, 
  Haokun Lin$^{2,3,4}$, 
  Yichen Wu$^{3,5}$,\\
  Shuyu Yang$^{1}$, 
  Zhongang Qi$^{6\ddagger}$,
  Chen Ma$^{3}$$^\ddagger$, 
  Li Zhu$^{1\ddagger}$, 
  and Ying Shan$^{2}$ \\
    {
  \normalsize
    $^{1}$Xi'an Jiaotong University\ \ \
    $^{2}$ARC Lab, Tencent PCG\ \ \
    $^{3}$City University of Hongkong\ \ \ 
    }\\
    {
    \normalsize
    $^{4}$Institute of Automation, CAS\ \ \
    $^{5}$Harvard University\ \ \
    $^{6}$vivo Mobile Communication Co. \ \ \
    }
  \thanks{
    Work is done during internship at Tencent.
  }
  \\
  \small $^*$Equal Contribution\hspace{0.2cm} $^\dagger$Project lead \hspace{0.2cm}
  $^\ddagger$Corresponding authors
  \hspace{0.1cm} 
\vspace{-1.0cm}
}

\begin{document}
\begin{CJK}{UTF8}{gbsn}
\maketitle

\begin{strip}
    \centering
  \includegraphics[width=0.9\textwidth]{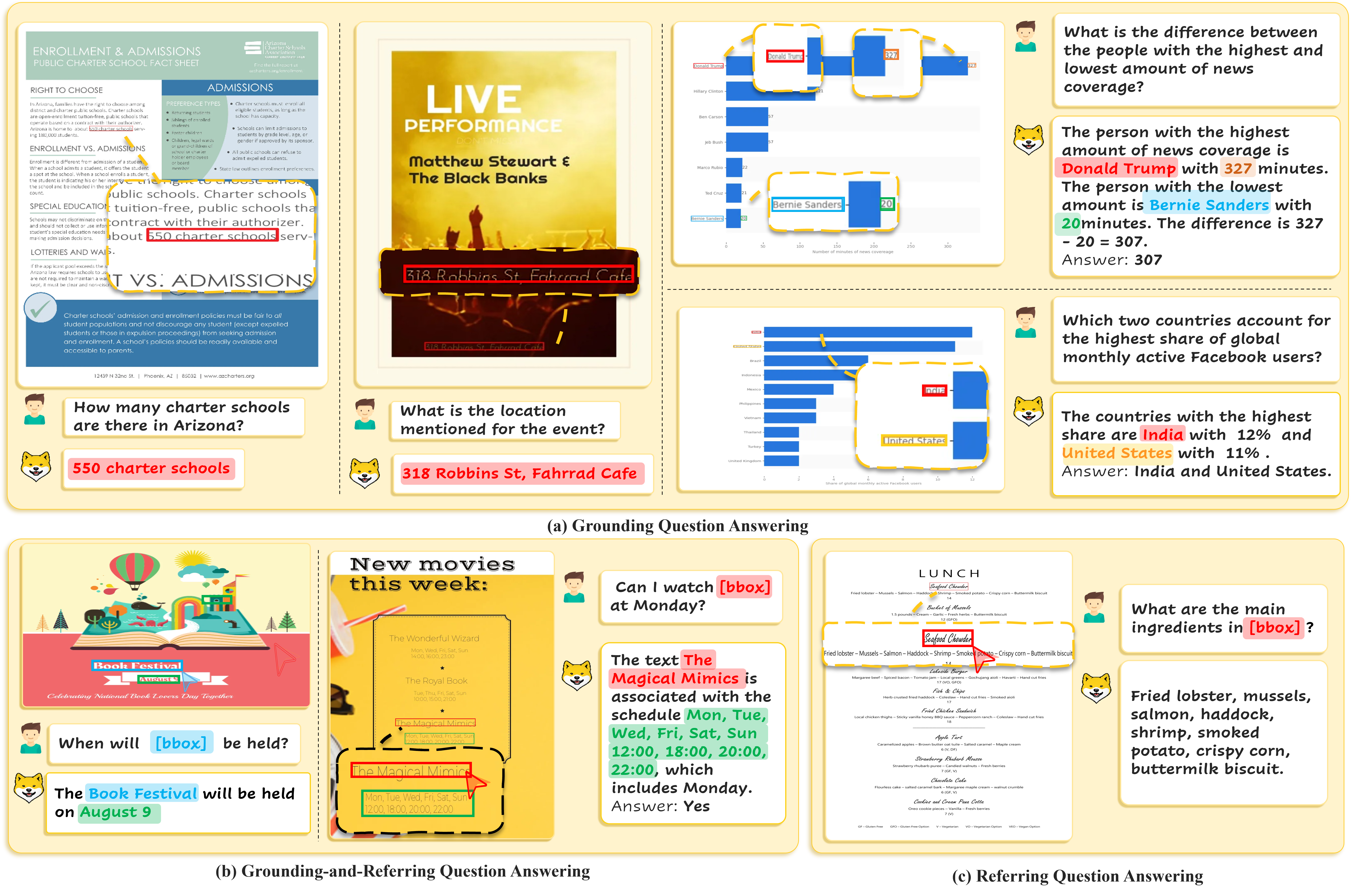}
    \vspace{-5mm}   
    \label{fig:teaser}
  \captionof{figure}{We propose DOGR, a multi-modal large language model that enables users to engage in versatile grounded document interactions. The figure illustrates the reasoning results of DOGR for grounding, grounding-and-referring, and referring tasks in DOGR-Bench.}
  \label{fig:teaser} 
    \vspace{-3mm}   
\end{strip}

\begin{abstract}
With recent advances in Multimodal Large Language Models (MLLMs), grounding and referring capabilities have gained increasing attention for achieving detailed understanding and flexible user interaction. However, these capabilities still remain underdeveloped in visual document understanding due to the scarcity of fine-grained datasets and comprehensive benchmarks.
To fill this gap, we propose the \textbf{\textit{DO}}cument \textbf{\textit{G}}rounding and \textbf{\textit{R}}eferring data engine (\textbf{\textit{DOGR-Engine}}), which generates two types of high-quality fine-grained document data: (1) multi-granular parsing data to improve text localization and recognition, and (2)
instruction-tuning data to activate MLLMs' grounding and referring capabilities in dialogue and reasoning. Using the DOGR-Engine, we construct \textbf{\textit{DOGR-Bench}}, a benchmark covering seven grounding and referring tasks across three document types (chart, poster, and PDF document), offering a comprehensive evaluation of fine-grained document understanding.
Leveraging the generated data, we further develop \textbf{\textit{DOGR}}, a strong baseline model that excels in text localization and recognition, while precisely grounds and refers to key textual information during conversation and reasoning, thereby advancing document understanding to a finer granularity and enable flexible interaction paradigms. Our code, data, and model are open-sourced at \url{https://github.com/zyinan99/DOGR}.

\end{abstract}
 
\section{Introduction}
\label{sec:intro}

Visual document understanding is challenging due to dense textual content and complex layouts, which hinder fine-grained comprehension.~Effective interaction with such documents requires the ability to refer to specific regions for precise understanding and accurately ground key details.

However, due to the lack of a clear definition for document grounding and referring tasks, as well as the absence of high-quality fine-grained data for training and evaluation, the advancement of multimodal large language models in this field has been notably limited. Many existing models~\cite{Qwen2.5-VL,chen2024expanding,gpt4o,gemini} struggle with basic fine-grained document tasks such as text localization and region-level recognition, let alone achieve effective grounding and referring to crucial textual elements during conversation or reasoning.

Several prior works either incorporate multi-granularity parsing data during pre-training to enhance document perception~\cite{mplug1.5} or introduce simple region-level instruction-tuning tasks to achieve basic referring capabilities~\cite{fox}.
While these methods aim to improve document understanding, they all suffer from two significant shortcomings:

\begin{itemize}
  \item[$\bullet$] {\bf Suboptimal annotation quality.} To collect parsing annotations, the Optical Character Recognition (OCR) tools are typically employed to extract text and bounding boxes.
  However, OCR tools encounter issues with inaccurate recognition of text and corresponding bounding boxes and have difficulty extracting semantically coherent text from documents with complex layouts, limiting their potential for developing fine-grained document understanding and question answering data.

  \item[$\bullet$] {\bf Lack of diversity in task formats.} Current instruction-tuning datasets primarily cater to basic referring tasks, such as region-level OCR, summarization and translation. However, these datasets fall short in supporting grounding tasks and fail to seamlessly incorporate grounding and referencing capabilities into the dialogue and reasoning processes of MLLMs. The restricted range of tasks hinders the model's ability to flexibly perceive details and affects the user interaction experience.
\end{itemize}

To solve the above two issues, we introduce the \textbf{DO}cument \textbf{G}rounding and r\textbf{E}ferring data engine (\textbf{DOGR-Engine}) for constructing high-quality fine-grained document data. DOGR-Engine generates: (1) 2.1M multi-granular document parsing data, which provides text box annotations at the word, phrase, line, paragraph and full-page level across posters, charts, and PDF documents. This dataset enhances text localization and recognition capabilities and serves as the foundation for creating instruction-tuning data. (2) A diverse set of 700K instruction-tuning samples, covering text-in location-out (grounding), location-in text-out (referring) data, and tasks combining location and text in both input and output. These data, generated from our multi-granular document parsing dataset with GPT-4o~\cite{gpt4o} assistance, ensure high linguistic quality and accurate grounding and referring annotations.

To benchmark the MLLMs' document grounding and referring capabilities comprehensively, as well as providing a clear task definition,

we propose \textbf{DOGR-Bench}, which contains 3.6K test samples and encompasses seven grounding and referring tasks across three document types (chart, poster, PDF document). On this benchmark, the state-of-the-art MLLMs demonstrate poor performance for grounding and referring tasks, suggesting that current models may not yet be equipped to handle the intricate challenges in fine-grained document understanding.
This performance gap underscores the DOGR-Bench’s value in identifying key limitations and guiding future advancements in MLLMs.

Furthermore, based on data generated by our engine, we develop a strong baseline model, \textbf{DOGR}, capable of understanding spatial referring and accurately grounding text
within document images. We report the performance of our model on DOGR-Bench, providing a performance reference for future research.

In summary, our contributions are threefold. (1) We introduce DOGR-Engine, a data construction pipeline that generates large-scale, high-quality multi-granular document parsing data and diverse ground-and-refer instruction tuning data. (2) We develop DOGR-Bench, the first comprehensive benchmark designed to evaluate MLLMs' grounding and referring capabilities in document understanding. (3) We present DOGR, a pioneering MLLM that is capable of understanding referred text and performing text grounding during conversation, leading to detailed document comprehension and more intuitive user interaction.

\begin{figure*}[t]
  \centering
  \includegraphics[width=0.9\linewidth]{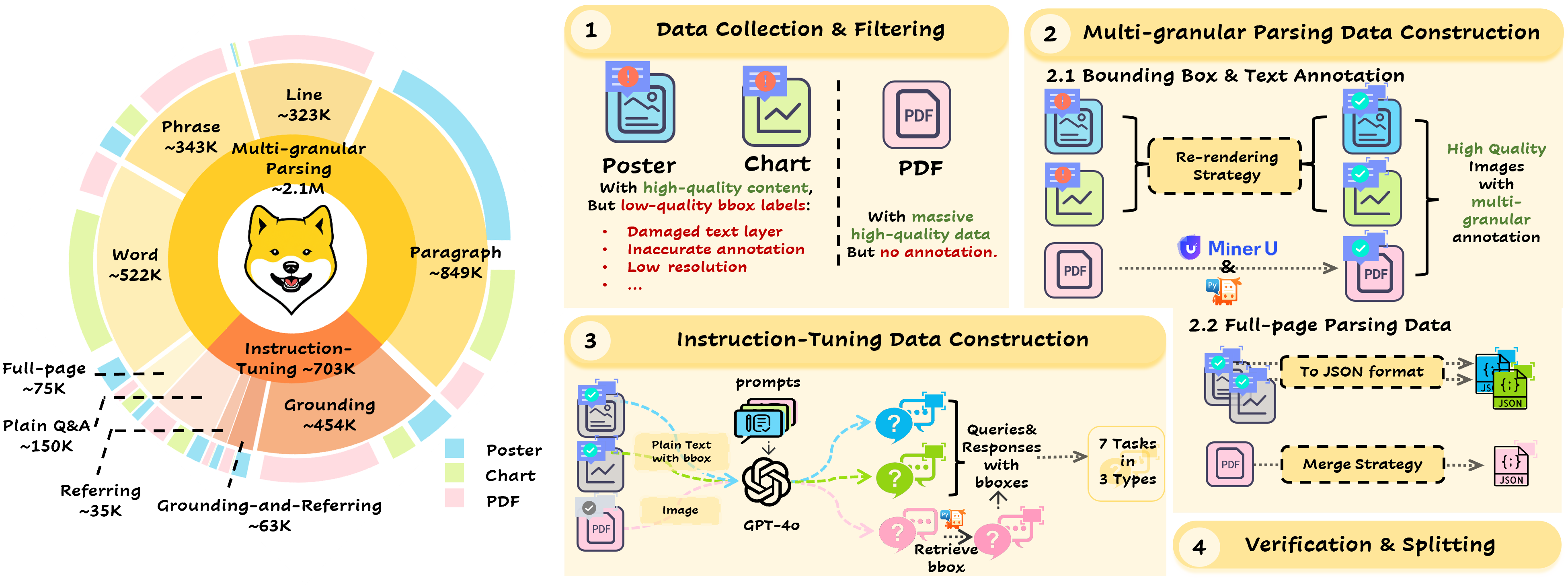}
   \vspace{-8pt}
   \caption{Left: Hierarchical task taxonomy and sample distribution analysis. Right: The pipeline of DOGR-Engine, which outlines the construction process for multi-granularity parsing data and ground-and-refer instruction-tuning data.
   }
   \label{fig:statistic}
   \vspace{-0.4cm}
\end{figure*}
\section{Related Work}
\label{sec:relatedwork}

\subsection{MLLMs for Visual Document Understanding}

Recently, several Multimodal Large Language Models~\cite{ye2023ureader,wei2023vary,internlmxcomposer2_4khd,chen2023internvl,zhou2025scale,lin2025toklip} have been introduced to perform visual document understanding without relying on OCR tools.
UReader~\cite{ye2023ureader} unifies a wide range of document understanding tasks with instruction-tuning format. A shape-adaptive cropping module is further designed to encode rich textual content in high-resolution image.
TextMonkey~\cite{liu2024textmonkeyocrfreelargemultimodal} employs shifted window to build connections between different image tiles, alleviating the issue of incoherence semantic caused by image cropping.
InternLM-XComposer-4KHD~\cite{internlmxcomposer2_4khd} and InternVL 1.5~\cite{chen2024far} further increase the tile number, significantly improving the performance on visual document understanding tasks.
These works achieve promising performance but lack document grounding and referring capabilities, which hinders the grounded document understanding and flexible human-AI interaction.

\subsection{MLLMs for Grounding and Referring}

In pursuit of fine-grained image understanding and convenient interaction, recent studies integrate grounding and referring abilities into MLLMs~\cite{chen2023shikra,kosmos-2,you2023ferret,zhang2025llava,Qwen2.5-VL,yang2024qwen2}. Kosmos2~\cite{kosmos-2}, Shikra~\cite{chen2023shikra} and Ferret~\cite{you2023ferret} utilize bounding boxes or visual prompts to pinpoint specific regions of an image and generate responses with key objects being grounded, facilitating flexible content referring and interaction. Additionally, LLaVA-Grounding~\cite{zhang2025llava} and GLaMM~\cite{hanoona2023GLaMM} further employ finer-grained multi-granularity masks for pixel-level grounding across various semantic levels.
These works perform well on real-world images but can't adapt to visual document understanding due to domain gaps.
Recently, there are several attempts to develop visual document grounding and referring. mPLUG-1.5~\cite{mplug1.5} and Kosmos-2.5\cite{lv2023kosmos} enhance localization and fine-grained perception with bounding box annotations from OCR tools and PDF parsing tools, respectively. Fox~\cite{liu2024focus} utilizes various visual prompts to refer to document regions and enables the MLLMs to extract or translate the region-level content. However, mPLUG-1.5 and Kosmos-2.5 only support basic text localization and recognition tasks. Fox further supports the referring translation task. These methods, though effective, 
fall short of integrating both grounding and referring into broader reasonings and dialogues. 
As a result, they leave the full potential of MLLMs for visual document grounding and referring largely untapped,
particularly in reasoning tasks where grounding plays a crucial role.

\section{DOGR-Engine}
\label{sec:dataset}
While well-annotated and diverse grounded data are crucial for improving the grounding and referring abilities of MLLMs~\citep{kosmos-2,tong2024cambrian,zhang2024mm1,lin2024drawandunderstand}, comprehensive and accurately labeled document-grounded datasets still remain scarce. Given the time-consuming and labor-intensive nature of manually annotating raw document images, we collect a large corpus of documents, including posters, charts, and PDF files, and develop \textbf{DOGR-Engine} to efficiently construct fine-grained document-grounded datasets. 

In this section, we first introduce the data processing pipeline and the generation of two types of high-quality document data.~Specifically, the data processing pipeline and the annotation process for multi-granular parsing data are described in \cref{sec:box}, while the construction pipeline for instruction-tuning data is elaborated in \cref{sec:qa}.~Additionally, an overview of statistical information is provided in \cref{sec:statistic}, and the annotation results are in \cref{sec:annotationresults}.

\subsection{Multi-granular Parsing Data Construction}
\label{sec:box}
As shown in \cref{fig:statistic}(Right), we start by filtering the raw data to remove low-quality samples or those with missing or broken information. We then describe the annotation process for multi-granular parsing data as follows.
\subsubsection{Automatic Bounding box Annotation}

\begin{figure*}[t]
  \centering

    \begin{subfigure}{0.44\linewidth}
    \raggedleft
    \includegraphics[width=1\linewidth]{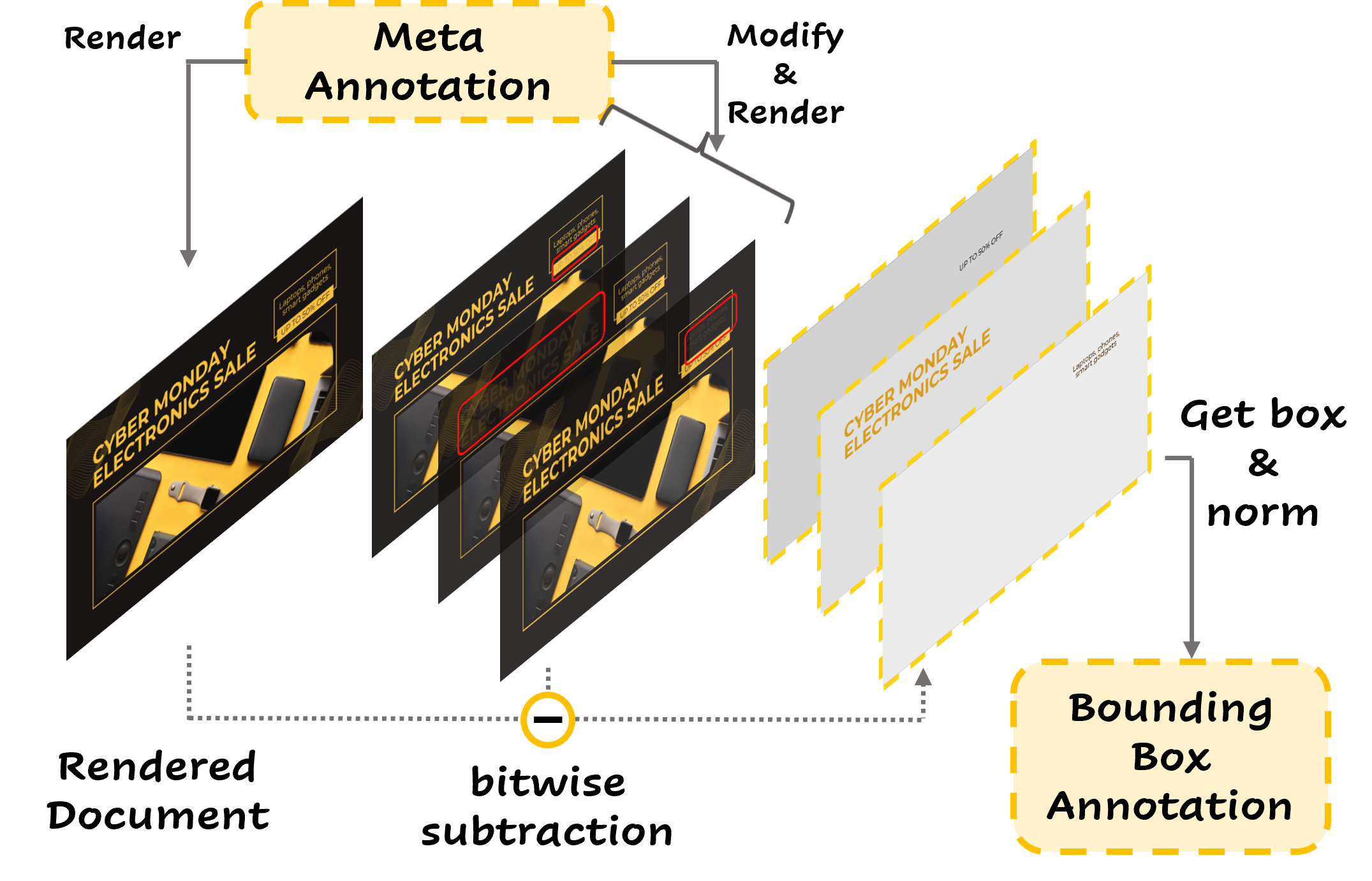}
    \caption{Re-rendering Strategy}
    \label{fig:bitwise}
  \end{subfigure}
  \hfill
  \begin{tikzpicture}     
    \draw [dashed] (0,-3.6) -- (0,2);
  \end{tikzpicture}
  \hfill
  \begin{subfigure}{0.52\linewidth}
    \raggedright
    \includegraphics[width=1\linewidth]{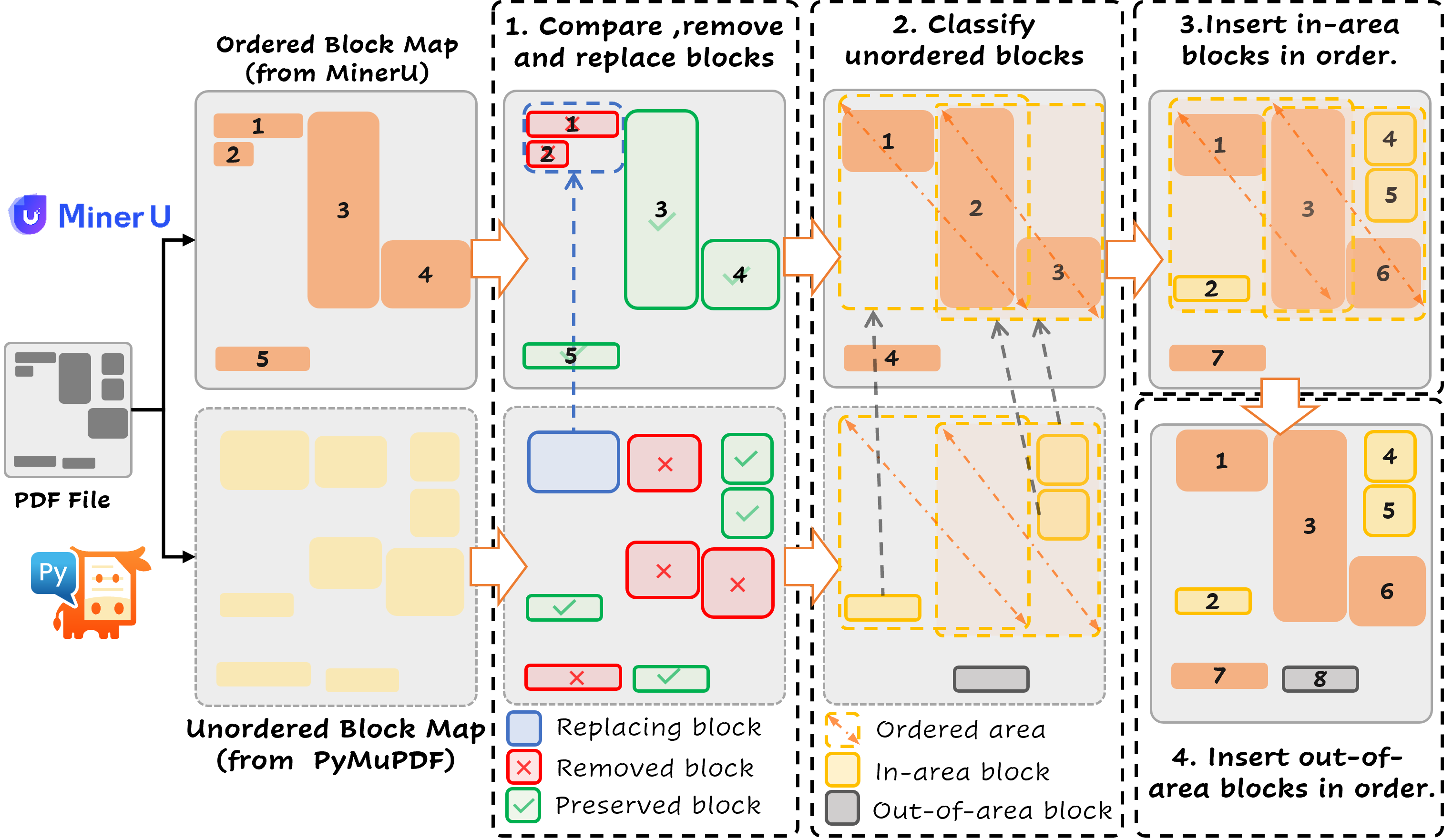}
    \caption{Merge Strategy}
    \label{fig:merge}
  \end{subfigure}
   \vspace{-10pt}
   \caption{(a) For the poster and chart data, we propose the Re-rendering Strategy to automatically obtain precise bounding boxes. (b) For the PDF document data, we propose the Merge Strategy combining the annotations from MinerU and PyMuPDF to obtain comprehensive and layout-aware full-page parsing annotations.}
   \label{fig:strategy}
   \vspace{-0.6cm}
\end{figure*}

\noindent{\bf Poster.}
We collect poster data from the Crello dataset~\cite{yamaguchi2021canvasvae}, which consists of poster templates from the design service website. Each poster contains the document meta-annotation, which includes rendered text blocks and their corresponding bounding boxes. However, we observe that some provided rendered text blocks are significantly degraded, and the bounding boxes are inaccurate. To address this issue, we re-render the text content and design a \textbf{Re-rendering Strategy} to obtain precise bounding boxes. Specifically, as illustrated in \cref{fig:bitwise}, the strategy consists of three steps: Firstly, we render the poster with the meta-annotation; Then, we modify the color or opacity attribute of one text block and perform a re-rendering process; Finally, since the rendering results of the first two steps are identical except for the modified attribute, the image of the target text block can be obtained by applying pixel-wise subtraction. By repeating the above steps for every text block, we can obtain accurate bounding boxes of all text blocks, which are then normalized to create the final annotation. Note that we preserve the original font format during re-rendering to keep the font diversity of poster data.

\noindent{\bf Chart.}
The chart data is collected from the ChartQA dataset~\cite{masry2022chartqa}, including various bar charts, line graphs, and pie charts, along with their corresponding JSON/CSV information. We extract the necessary information for rendering images and utilize Matplotlib to create chart images that closely resemble the original charts. To increase the position diversity of text blocks, we apply random padding around the edges of the chart images. Additionally, to prevent the model from over-relying on text and neglecting its ability to interpret visual elements, e.g., bars and lines, we propose to remove the text data from one-third of the chart data, and randomly mask half of the text in another third. Finally, we reuse the Re-rendering Strategy to obtain bounding box annotations for all text blocks, including the values, titles, legends, and axis labels.

\noindent{\bf PDF Document.} 
We select a high-quality subset from CC-MAIN-2021-31-PDF-UNTRUNCATED~\cite{CC-MAIN-2021-31-PDF-UNTRUNCATED}, which consists of a large collection of various text documents in PDF format. Leveraging the PDF parsing tool PyMuPDF~\cite{PyMuPDF} and the document content extraction framework MinerU~\cite{wang2024mineruopensourcesolutionprecise}, we directly extract parsing data of different granularities (word, phrase, line, paragraph).

\subsubsection{Full-page Parsing Data Construction}
\label{sec:fullpage}
In addition to the detailed annotations across word, phrase, line, and paragraph-level, we also construct full-page parsing annotations to enhance the comprehensive perception of the document content.

\noindent \textbf{Poster and Chart.} The poster data features sparse textual content, making it straightforward to grasp the dependencies between text blocks. Therefore, we organize the paragraph-level text and their corresponding bounding boxes in a left-to-right, top-to-bottom scanning order to create the full-page parsing annotation in JSON format. In contrast, given the highly organized nature of the chart data, we structure its full-page parsing annotations as a JSON dictionary containing hierarchically organized components: chart title, axis labels, legends, and data markers—each precisely associated with their spatial coordinates.

\noindent {\bf PDF Document.} PDF document contains extensive text and complex layouts, necessitating parsing in a logical reading order to comprehend the information accurately. MinerU~\cite{wang2024mineruopensourcesolutionprecise} contains a layout detection model and an OCR model, allowing for extracting text blocks with a certain reading order. However, it often fails to capture all content in documents, missing tables, footnotes, and other elements. In contrast, PyMuPDF~\cite{PyMuPDF} can thoroughly extract content from documents, but it does not provide an appropriate reading order. Therefore, we propose a \textbf{Merge Strategy} that combines the strengths of MinerU and PyMuPDF to achieve comprehensive, layout-aware annotations. As shown in \cref{fig:merge}, 1) we first compare text blocks from ordered and unordered maps to eliminate duplicate text blocks. For the truncated blocks (red block 1 and red block 2 in the ordered map), we replace them with the corresponding complete block (blue block in the unordered map) to improve the semantic completeness within blocks; 2) In the ordered map, we construct an ordered area if two consecutive blocks are placed from the top left to the bottom right. We classify the preserved blocks in the unordered map into two categories: in-area blocks and out-of-area blocks; 3) We insert the in-area blocks into the ordered area and sequentially update the block order in each ordered area using column-major order; 4) For the out-of-area blocks,  we insert them into the ordered map. The order of these blocks is then determined based on their positional relationship with their nearest ordered blocks, following column-major order.

\subsection{Instruction-Tuning Data Construction}
\label{sec:qa}
\vspace{-0.2cm}
To seamlessly integrate grounding and referring capabilities in the dialogue and reasoning, it is crucial to construct high-quality instruction-tuning data with accurate grounded text in both query and response. Based on the precise bounding box annotations from \cref{sec:box}, we leverage GPT-4o~\cite{gpt4o} to generate diverse-formatted instruction-tuning data tailored to various text granularities, encompassing tasks such as question answering, reasoning, and summarization. 
To further enhance instruction-following quality, we first constrain the length and type of the generated responses, resulting in different response categories (e.g., short responses, open-ended long responses).~Then, we incorporate a response format prompt into each query. The detailed
prompts for GPT-4o are shown in \cref{sec:prompts}

\noindent
{\bf Poster and Chart.} As introduced in \cref{sec:fullpage}, the full-page parsing annotation for poster and chart data includes all text blocks along with their corresponding bounding boxes, which contain sufficient information to construct ground-and-refer instruction-tuning data.
Therefore, we provide GPT-4o with the full-page parsing data, and task GPT-4o to generate queries and responses in a grounded manner, where the generated content must include texts that originate from the given parsing annotations. 
Additionally, we require GPT-4o to wrap the texts originating from parsing annotations with `\texttt{<ocr></ocr>}' and append the corresponding coordinates wrapped in `\texttt{<bbox></bbox>}'.

\noindent
{\bf PDF Document.}  
PDF documents often contain extensive text content. If we input all the text of various granularities into GPT-4o, it would result in excessively long prompts. This not only incurs significant API calling costs but also increases the difficulty for GPT-4o in comprehending and completing the tasks, leading to low generation quality. To address these issues, we design a {\bf Post-annotating Strategy} to generate high-quality data with minimal overhead. Specifically, 1) We first use the document image as input for GPT-4o instead of the text of the document, which significantly reduces the token count; 2) Then, we task GPT-4o to generate queries and responses based on the document images. Additionally, any text within the queries and responses that originates from the document image must be wrapped with ``\texttt{<ocr></ocr>}''; 3) We extract the texts wrapped with ``\texttt{<ocr></ocr>}'' and utilize PyMuPDF to retrieve the corresponding bounding boxes; 4) Finally, we wrap these bounding boxes with `\texttt{<bbox></bbox>}' and insert them back into the generated content after their corresponding texts. For texts that cannot be located by PyMuPDF, we remove wrapped tokens and convert them to plain text. This approach allows the generation of diverse instruction-tuning data at a low cost while ensuring the quality of grounding and referring annotations.

\subsection{Data Verification and Splitting}
\label{sec:statistic}

Although we require GPT-4o to generate grounded responses in the format of ``\texttt{<ocr> text </ocr> <bbox> x1, y1 ,x2 ,y2 </bbox>}'',~GPT-4o sometimes fails to follow our requirement, resulting in wrong coordinates format or missing ``\texttt{<bbox></bbox>}''. Therefore, we implement a rule-based filter
to remove these defective samples.
Additionally, for poster and chart data, we extract the grounded text from the generated content and compare them with the full-page parsing annotations, filtering out samples that contain incorrect grounded text. 
Furthermore, we also performed a detailed categorization of the data.
As shown in \cref{fig:statistic}(Left), DOGR-Dataset includes 2.1M multi-granular parsing data and 700K diverse-formatted instruction-tuning data across three document types: poster, chart, and PDF document. The multi-granular parsing data includes four fine-grained levels (word, phrase, line, paragraph) and a full-page level, with precise bounding box annotations for the grounded text at different granularities.
The instruction-tuning data comprises four types: grounding, referring, grounding-and-referring, and plain Q\&A, with the plain Q\&A subset derived by removing the grounded content from a portion of the grounding data. This subset is created to enhance data diversity while preserving traditional document understanding capabilities.
More dataset statistics and construction details are shown in \cref{sec:Datasetdetails}.

\section{DOGR-Bench}
\label{sec:bench}

To evaluate the grounding and referring capabilities of MLLMs on visual document understanding tasks, we introduce the DOGR-Bench in this section.

\begin{figure}[t]
  \centering
   \vspace{-0.3cm}
  
  \includegraphics[width=1\linewidth]{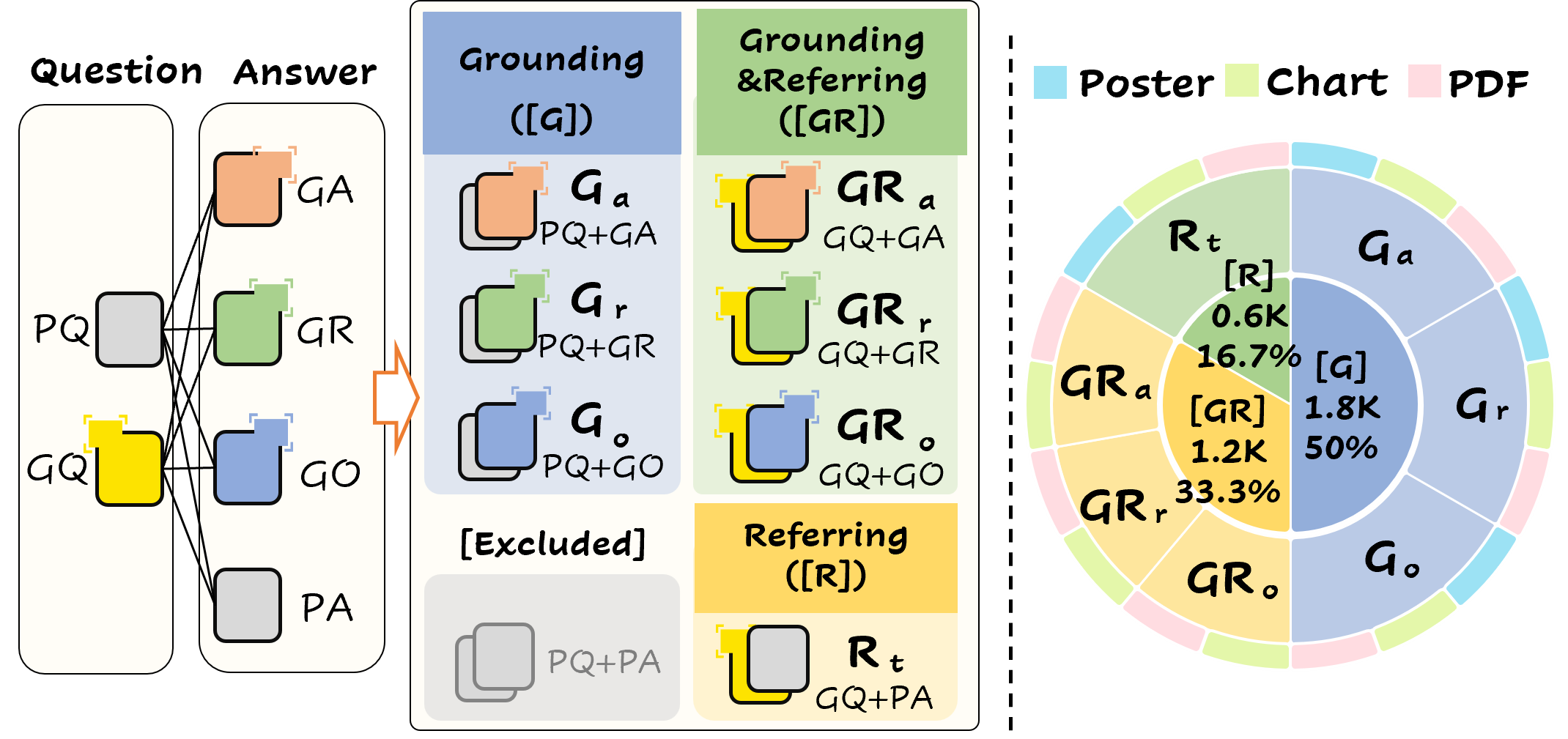}
   \caption{Left: In DOGR-Bench, we categorize the samples into 7 evaluation tasks. Right: Data statistics of DOGR-Bench.}
   \label{fig:bench_construct}
    
   \vspace{-0.6cm}
\end{figure}

\noindent
\textbf{Task Definition.} 
As shown in~\cref{fig:bench_construct}~(Left), we systematically construct our benchmark by categorizing our data into distinct classes based on both input and output formats. This classification helps in designing clear evaluation metrics. 
We divide the input formats into two categories based on the presence of bounding boxes: \textbf{Grounded Question (GQ)} with bounding boxes, and \textbf{Plain-Text Question (PQ)} without bounding boxes.
The output formats are categorized into four classes:

\begin{itemize} \item \textbf{Grounded Answer(GA)}: 
The response consists of a brief answer accompanied by its corresponding bounding box.

\item \textbf{Grounded Reasoning(GR)}: The response includes the detailed reasoning process and the final answer, while the key text contents in the reasoning process are grounded.


\item \textbf{Grounded Open-ended Answer(GO)}: An open-ended response with one or more key text contents grounded, without providing an answer in a certain format.

\item \textbf{Plain Text Answer(PA)}: This format does not incorporate grounded text content. \end{itemize}
By combining two input forms and four output forms, we derive 7 document grounding and referring tasks. Among these tasks, 3 tasks primarily assess grounding capability: grounded answering for plain text questions (${G_a}$), grounded reasoning for plain-text questions (${G_r}$), and grounded open-ended answering for plain-text questions (${G_o}$). The plain-text answering for grounded questions (${R_t}$) task evaluates referring capability. The remaining tasks, grounded answering with plain-text questions (${GR_a}$), grounded reasoning for grounded questions (${GR_r}$), and grounded open-ended answering for grounded questions (${GR_o}$) require the integration of both grounding and referring capabilities for successful completion.

\noindent
\textbf{Metrics.} 
Our benchmark evaluation encompasses two aspects: grounding performance and text answer accuracy. 
Following the previous works in grounded captioning~\cite{you2023ferret}, we evaluate the grounding and text answer separately.
For grounding performance, we use $F1_{all}$ score, which evaluates grounding results as a multi-label classification problem. The generated grounded text is considered correct if the Intersection over Union (IoU) between its bounding box and the GT bounding box is greater than 0.5, meanwhile its text matches with the GT text. 
For text answer accuracy, we use exact text matching accuracy for short-answer tasks and BLEU score for long-answer tasks.

\noindent \textbf{Data Statistics.} DOGR-Bench includes 1.8k grounding, 0.6k referring, and 1.2k grounding-and-referring samples. See detailed statistics in \cref{fig:bench_construct}~(Right).

\section{DOGR}
\label{sec:model}
\begin{figure}[t]
  \centering
  \includegraphics[width=0.98\linewidth]{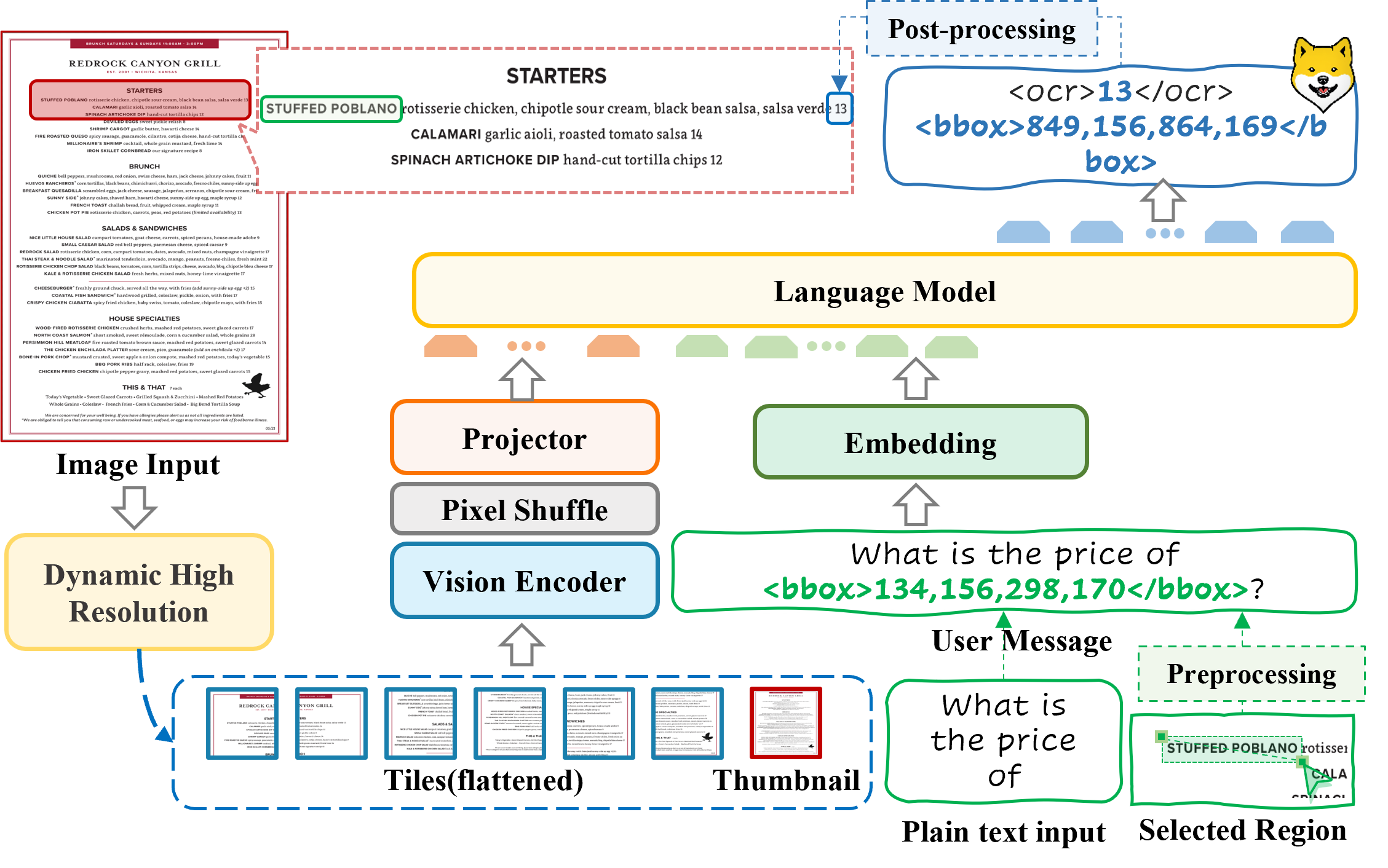}
   \vspace{-10pt}
   \caption{The overall architecture of DOGR.}
   \label{fig:model}
   \vspace{-0.7cm}
\end{figure}

\noindent \textbf{Overall Architecture.} As illustrated in \cref{fig:model}, our model DOGR employ a general MLLM architecture, including a vision encoder, a projector, and a large language model. In the visual encoder component, to enable the model to handle high resolution, we first search for the best aspect ratio for the input image and dynamically segment images into multiple tiles. These tiles, along with a thumbnail of the input image, are provided as input to the vision encoder. 
We employ pixel shuffle\cite{chen2024far} to improve the computational efficiency of the model when processing high-resolution images. 
For bounding boxes, we simply discretize the continuous coordinates into discrete values from 0 to 999, avoiding the introduction of extra modules or location tokens.
During inference, we transfer the user-selected regions to coordinates of bounding boxes and insert them into the query for preprocessing. After obtaining the output, we utilize post-processing to overlay bounding boxes on original document images, thereby facilitating user interaction.

\noindent \textbf{Training Strategy.} 
We adopt a three-stage training strategy, including pre-aligning, pre-training, and fine-tuning. The pre-aligning stage focuses on aligning the feature space of vision and language rapidly. We freeze both the vision encoder and the large language model, and train the projector with a relatively large learning rate. The pre-training stage aims at document parsing capabilities. We unfreeze the vision encoder and the LLM, enabling the model to recognize diverse textual content and acquire text-reading capability. In the fine-tuning stage, we train the entire model using diverse instruction-tuning data, enhancing its instruction-following ability while activating its grounding and referring capabilities.

\noindent \textbf{Training Dataset.} In the pre-aligning stage, we utilize LLaVA-558K \cite{llava} to train the projector. For the pre-training dataset, we utilize DocStruct4M~\cite{mplug1.5} along with our 2.1M multi-granular parsing data to enhance basic text reading, text grounding and text referring capabilities of DOGR. For the fine-tuning data, we adopt our ground-and-refer instruction-tuning data and meticulously selected datasets to enhance model performance across a wide range of document parsing and understanding tasks, resulting in a final fine-tuning dataset of ~2M samples. The detailed composition of fine-tuning dataset can be found in supplementary materials.
In \cref{sec:Trainingdetails}, we provide a detailed description and statistics of our training data composition.

\vspace{-0.1cm}
\vspace{-0.1cm}
\section{Experiment}
\vspace{-0.1cm}
\label{sec:experiment}

\begin{table*}[h]
\centering
\vspace{-15pt}

\caption{Performance comparison on DocLocal4K. $\dagger$ denotes the pre-trained model using parsing data with bounding boxes.}
\vspace{-4pt}
\label{tab:pretrain}
\resizebox{0.85\linewidth}{!}{
    
    \begin{tabular}{l|ccccc|ccccc}
    \toprule
    \multirow{2}{*}{{\textbf{Model}}}  & \multicolumn{5}{c|}{\textbf{Text Localization}} &
    \multicolumn{5}{c}{\textbf{Text Recognition}} \\ 
     \cmidrule(lr){2-6}\cmidrule(lr){7-11}
    
     & word&phrase&line&paragraph&ALL& word&phrase&line&paragraph&ALL \\
     \midrule
     Gemini 1.5 Pro\cite{gemini}&6.55&3.69&4.37&5.98&5.18&4.24&2.58&3.41&16.12&11.92\\
     Gemini 2.0 Flash&4.03&5.54&7.36&9.69&6.49&6.18&2.74&3.06&1.01&2.17\\
     Gemini 2.5 Flash\cite{comanici2025gemini}&8.57 & 8.86&10.74 & 10.72&9.65 &14.51&12.45&17.41&37.71&25.04\\
     Gemini 2.5 Pro\cite{comanici2025gemini}&23.19 &39.11 &38.57 &10.10 &27.91 & 37.22&32.09 &45.39 &47.90 &46.27\\
     GPT-4o\cite{gpt4o}&6.89 & 2.77& 3.18&8.25 &5.27 & 6.02&2.81 & 4.04&3.86 & 4.38\\
     InternVL2.5-8B\cite{chen2024expanding}& 8.89&5.19 & 2.60& 12.61& 7.21&6.88 &3.74 & 4.65& 9.08& 8.29\\
     Qwen2.5-VL-7B \cite{Qwen2.5-VL}& 20.84& 13.84&25.25 & 27.01& 21.51&28.68&14.82& 22.5& 30.51&28.98\\

    DocOwl-1.5$\dagger$\cite{mplug1.5}&70.42 &76.38  &85.88 & 91.34 &80.38 &
    70.10& 67.86 & 73.88& 70.70 &70.63 \\
     \midrule

    \rowcolor{yellow!30}
    \textbf{DOGR$\dagger$} & \textbf{81.85}& \textbf{83.58} & \textbf{86.88}& \textbf{95.67}&\textbf{86.64}& \textbf{80.00}& \textbf{78.85} & \textbf{77.60}& \textbf{72.72} &\textbf{77.88} \\
    \bottomrule
    \end{tabular}
}
\end{table*}

\begin{table*}[h]
    \centering
    \vspace{-5pt}
    \caption{ Comparison of existing MLLMs' and DOGR's performance on DOGR-Bench. For clear comparison, we relax the IoU threshold to 0.1 when evaluating ${F1_{all}}$(indicated with *). In the final results for DOGR, we set the IoU threshold to 0.5.}
    \label{tab:morebenchresult}
    \vspace{-4pt}
    \resizebox{0.85\linewidth}{!}{\begin{tabular}
    {l|cccccc|cccccc|c}
    \toprule
    &  \multicolumn{6}{c|}{\textbf{Grounding}} & \multicolumn{6}{c|}{\textbf{Grounding-and-Referring}} & \textbf{Referring}\\
     \cmidrule(lr){2-7}\cmidrule(lr){8-13}\cmidrule(lr){14-14}
         \multicolumn{1}{l|}{\textbf{ Model}}&   \multicolumn{2}{c|}{${G_a}$} &  \multicolumn{2}{c|}{${G_r}$}  &  \multicolumn{2}{c|}{${G_o}$}
         &  \multicolumn{2}{c|}{${GR_a}$}  & \multicolumn{2}{c|}{${GR_r}$} &\multicolumn{2}{c|}{${GR_o}$} &${R_t}$\\
         \cmidrule(lr){2-7}\cmidrule(lr){8-13}\cmidrule(lr){14-14}
         & \multicolumn{1}{c|}{${Acc}$}& \multicolumn{1}{c|}{${F1_{all}}$}& \multicolumn{1}{c|}{${Acc}$}& \multicolumn{1}{c|}{${F1_{all}}$}&\multicolumn{1}{c|}{BLEU4}& \multicolumn{1}{c|}{${F1_{all}}$}&\multicolumn{1}{c|}{${Acc}$} &\multicolumn{1}{c|}{${F1_{all}}$} & \multicolumn{1}{c|}{${Acc}$}& \multicolumn{1}{c|}{${F1_{all}}$}&\multicolumn{1}{c|}{BLEU4}& \multicolumn{1}{c|}{${F1_{all}}$}& \multicolumn{1}{c}{${Acc}$}\\ \midrule
       Gemini 1.5 Flash*\cite{gemini}  &32.7 &0.7&59.0&1.1&13.6&1.7&42.8&0.5&31.8&0.3&13.5&0.1&37.5\\
       GPT-4o mini*\cite{gpt4o}  &64.5 &0.8&48.7&0.6&9.4&0.5&40.5&0.0&23.3&0.7&7.5&0.2&18.8\\
       Gemini 1.5 Pro*\cite{gemini}  & 77.7&9.4&62.0&5.8&13.4&6.7&46.0&5.7&29.8&2.5&17.4&3.7&37.2\\
       Gemini 2.0 Flash*  & 74.2&5.2&63.5&3.1&15.2&2.9&30.0&0.5&28.5&1.7&26.2&0.6&37.5\\
       Gemini 2.5 Flash*\cite{comanici2025gemini}  & 80.2&38.8&59.3&25.4&12.8&21.5&55.0&22.8&42.0&14.7&16.1&8.6&40.7\\
       Gemini 2.5 Pro*\cite{comanici2025gemini}  & 61.8&30.1&47.8&17.2&13.6&14.7&44.3&17.3&36.0&10.4&19.8&8.7&36.2\\
       GPT-4o*\cite{gpt4o}  & 79.0&8.8&47.0&3.8&12.5&9.2&50.0&0.5&26.8&0.8&12.0&1.2&39.8\\
       
       Qwen2-VL-7B*\cite{Qwen2-VL}  & 41.7&1.8&4.3&1.4&9.8&5.5&11.8&0.3&0.5&1.5&10.1&0.0&35.5\\
       
       InternVL2-8B*\cite{chen2024far}  &51.2 &2.4&26.2&0.1&10.2&1.1&26.3&0.0&12.8&0.0&14.5&0.0&17.7\\
         
       InternVL2.5-8B*\cite{chen2024expanding}  & 58.3&3.8&33.0&0.6&13.8&2.0&23.3&0.0&15.5&0.0&18.0&0.3&8.5\\
       
       Qwen2.5-VL-7B*\cite{Qwen2.5-VL}  & 63.8&19.5& 35.0&9.8 & 6.1&11.3 &40.0&1.5& 19.0& 4.35& 8.1& 0.6&43.0\\

        DOGR* & 83.2 & 76.3 &67.7 & 54.8 & 38.3 & 59.4 & 82.8 & 71.4 &68.0 &39.9 &43.0 &37.6&60.3\\
       \midrule
        
       \rowcolor{yellow!30}
        \textbf{DOGR} & \textbf{83.2} & \textbf{73.0} & \textbf{67.7} &\textbf{ 52.5} & \textbf{38.3 }& \textbf{57.1} & \textbf{82.8} & \textbf{66.9} &\textbf{68.0} & \textbf{38.2}& \textbf{43.0}&\textbf{34.9}&\textbf{60.3}\\
       \rowcolor{posterblue!20}
        ~~~-~Poster  & 79.5 &  75.8& 57.0 & 42.6 & 45.6 & 58.9 & - & - &- &-& -&-& 73.5\\
        \rowcolor{chartgreen!20}
        ~~~-~Chart & 90.5 &72.4 & 69.5 &56.9 & 32.3 & 62.8 & 89.0 & 74.5 &70.0& 39.1& 45.2&50.7 &60.0\\
        \rowcolor{pdfred!20}
        ~~~-~PDF Document & 79.5 & 70.7 & 76.5 &57.8 & 36.8 & 49.8 & 76.5& 59.5 & 66.0&37.3&40.8 &19.0&47.5\\

        \bottomrule
    \end{tabular}}
        \vspace{-13pt}

\end{table*}

\begin{table*}[t]
\vspace{-12pt}
\caption{Performance comparison on 10 general document benchmarks.}
\vspace{-8pt}
\centering

\resizebox{0.85\linewidth}{!}{
\begin{tabular}{ll|cccc|cc|c|cc|c}
\toprule

\textbf{Model} & 
\textbf{Size}  &
{\begin{tabular}[c]{@{}c@{}}\textbf{Doc}\\\textbf{VQA}\end{tabular}} & 
{\begin{tabular}[c]{@{}c@{}}\textbf{Info}\\\textbf{VQA}\end{tabular}} & 
{\begin{tabular}[c]{@{}c@{}}\textbf{Deep}\\\textbf{Form}\end{tabular}}& 
{\textbf{KLC}} & {\textbf{WTQ}} & 
{\begin{tabular}[c]{@{}c@{}}\textbf{Tab}\\\textbf{Fact}\end{tabular}}&
{\begin{tabular}[c]{@{}c@{}}\textbf{Chart}\\\textbf{QA}\end{tabular}}&
{\begin{tabular}[c]{@{}c@{}}\textbf{Text}\\\textbf{VQA}\end{tabular}}&
{\begin{tabular}[c]{@{}c@{}}\textbf{Text}\\\textbf{Caps}\end{tabular}}&
{\begin{tabular}[c]{@{}c@{}}\textbf{Visual}\\\textbf{MRC}\end{tabular}}
\\ \midrule
IXC 2.5-7B\cite{internlmxcomposer2_5}&  7B    &   90.9 &  70.0  & \textbf{71.2} & - &\underline{53.6}&\textbf{85.2}&82.2&\underline{78.2}&-&\underline{307.5}\\ 
DocOwl-1.5-Chat\cite{mplug1.5}&  8B    &   82.2 &  50.7  & 68.8 & \underline{38.7} &40.6&80.2&70.2& 68.6&131.6&246.4\\ 
DocOwl-2\cite{mplug2}&  8B    &   80.7 &  46.4  & 66.8 & 37.5 &36.5&78.2&70.0& 66.7&\underline{131.8}&217.4\\
InternVL2-8B\cite{chen2024far}&  8B    &   91.6 &  \underline{74.8}  & -  &-& -&-&\underline{83.3}& 77.4&-&-\\ 
Qwen2-VL-7B\cite{Qwen2-VL}&  8B    &  \textbf{94.5} & \textbf{76.5}  & -  &-& -&-&83.0& \textbf{84.3}&-&-\\ 

\midrule
\rowcolor{yellow!20}
\textbf{DOGR}&  8B    & \underline{91.7} &{70.7}  & \underline{70.8}& \textbf{40.4} &\textbf{58.8}& \underline{84.5}&\textbf{83.6}&76.6& \textbf{145.9}&\textbf{332.5}\\ 
\bottomrule
\end{tabular}
}
\label{tab:maintable}
\vspace{-0.5cm}
\end{table*}

\begin{table}[h]
\centering
\caption{Our multi-granular parsing data ${\mathrm{MG}}$ enhances the model's basic document localization capability.}
\vspace{-5pt}
\label{tab:stage1_abl2}
\resizebox{1\linewidth}{!}{
    
    \begin{tabular}{l|cccc|cccc}
    \toprule
    \multirow{2}{*}{{\textbf{Model}}}  & \multicolumn{4}{c|}{\textbf{Text Localization}} &
    \multicolumn{4}{c}{\textbf{Text Recognition}} \\ 
     \cmidrule(lr){2-5}\cmidrule(lr){6-9}
    
     & word&phrase&line&paragraph& word&phrase&line&paragraph \\
     \midrule
    
    baseline &74.79& 78.41 & 84.1&95.46  &
   79.87& 75.73 &73.88 & 70.10  \\
    \rowcolor{yellow!30}
    ~\textbf{+ ${\mathrm{MG}}$} & \textbf{81.85}& \textbf{83.58} & \textbf{86.88}& \textbf{95.67}& \textbf{80.00}& \textbf{78.85} & \textbf{77.60}& \textbf{72.72}  \\
    \bottomrule
    \end{tabular}
    \vspace{-15pt}
}
\end{table}
\begin{table}[t]
\vspace{-8pt}

\caption{Our instruction-tuning data $\mathrm{{IT}}$ improves the model's traditional document understanding performance.}
\vspace{-5pt}
\centering

\resizebox{1\linewidth}{!}{
\begin{tabular}{l|cccc|c|c}
\toprule

\textbf{Model} & 
{\begin{tabular}[c]{@{}c@{}}\textbf{Doc}\\\textbf{VQA}\end{tabular}} & 
{\begin{tabular}[c]{@{}c@{}}\textbf{Info}\\\textbf{VQA}\end{tabular}} & 
{\begin{tabular}[c]{@{}c@{}}\textbf{Deep}\\\textbf{Form}\end{tabular}}& 
{\textbf{KLC}}& 
{\begin{tabular}[c]{@{}c@{}}\textbf{Chart}\\\textbf{QA}\end{tabular}}&
{\begin{tabular}[c]{@{}c@{}}\textbf{Visual}\\\textbf{MRC}\end{tabular}}
\\ \midrule

{baseline}   &   87.57 &  64.58 & 66.13 & 37.10 &80.88&265.9\\ 

\rowcolor{yellow!20}
\textbf{~+ ${\mathrm{IT}}$}  & \textbf{89.24} &\textbf{67.45}  & \textbf{68.89}& \textbf{38.32} &\textbf{81.92}&\textbf{287.25}\\ 
\bottomrule
\end{tabular}
}
\label{tab:baseline}
\vspace{-15pt}
\end{table}

\subsection{Implementation Details}
\label{sec:Implementation}

DOGR utilizes the InternViT-300M-448px \cite{chen2023internvl,chen2024far} as the vision encoder and Qwen2-7B-Instruct \cite{yang2024qwen2} as the LLM.
In the pre-aligning stage, we only train the projector and the learning rate is set to 1e-3. In the pre-training and fine-tuning stage, all model parameters are trainable. The learning rate for the vision encoder is 2e-6, while the learning rate for other components is 1e-5. Each stage is conducted for 1 epoch. In the pre-aligning stage, the batch size is set to 256, and we only use the thumbnail for vision encoder input. In the pre-training stage, the batch size is configured to 512. The number of image tiles and the max length of the input sequence to the large language model are set to 9 and 4096, respectively.
In the fine-tuning stage, the batch size is adjusted to 256, and the image tile number and max input length increase to 16 and 6144, respectively.

\subsection{Doc Grounding \& Referring Evaluation}
\label{sec:grounding}
\noindent
\textbf{Basic capability verification.}
We first benchmark existing MLLMs on DocLocal4K\cite{mplug1.5} to evaluate basic text localization and recognition abilities. We report IoU@0.5 for text localization and BLEU-4 score for text recognition at word, phrase, line, and paragraph levels. As quantified in \cref{tab:pretrain}, current SOTA MLLMs\cite{gemini,gpt4o,internlmxcomposer2_5,Qwen2.5-VL} show poor performance on both tasks. 
This underscores the lack of basic capabilities of existing MLLMs in fine-grained document tasks, \ie, detailed text position perception and region-level text recognition. This lack of basic capabilities also limits the model's ability to conduct further flexible grounding and referring.
On the contrary, pre-training with document parsing data endows DOGR and DocOwl-1.5\cite{hu2024docowl} with basic capabilities for text localization and recognition. Moreover, our high-quality and rich document parsing data further enhances the performance of DOGR.

\noindent
{\bf Performance comparison on DOGR-Bench.}
As shown in \cref{tab:morebenchresult}, our comprehensive evaluation reveals distinct performance patterns among existing MLLMs\cite{gpt4o,gemini,chen2024far,chen2024expanding,Qwen2-VL,Qwen2.5-VL} on DOGR-Bench. Notably, models such as GPT-4o (79.0\% Ga $Acc$) and Gemini 1.5 Pro (77.7\% Ga $Acc$) demonstrate relatively strong performance in basic document understanding tasks, indicating that these models possess fundamental document understanding abilities. However, correspondingly, the lower $Acc$ of referring-related tasks reveals their limitations in accurately perceiving and understanding bounding box coordinates.
More crucially, even IoU threshold for $F1_{all}$ is relaxed to 0.1, the other MLLMs' $F1_{all}$ scores across all tasks are consistently low, which indicates deficiencies in precise document grounding. In contrast, our model achieves superior performance in both understanding and grounding capabilities. Most remarkably, its grounding capability demonstrates a substantial advancement, substantially outperforming all existing open-source and proprietary SOTA models across multiple evaluation metrics.
The detailed evaluation process and results are shown in \cref{sec:Experiments}.

\noindent
{\bf Detailed Performance of DOGR.}
In the last 4 rows of \cref{tab:morebenchresult}, we detail the performance of DOGR across 3 data types and 7 tasks. Posters have a diverse range of font types and colors, making them ideal for assessing a model's adaptability to different font styles.
Charts contain fine-grained and structural elements, necessitating precise text localization and understanding capabilities. PDF document data typically features high-resolution and complex content, posing challenges on both referring and grounding tasks. Despite these challenges, DOGR demonstrates strong performance, validating its effectiveness and robustness.
More quantitative results are shown in \cref{sec:qualitative}.

\subsection{Traditional Document Understanding}
\label{sec:general}
To assess the overall capability of DOGR, we conduct experiments on 10 traditional document understanding benchmarks, 
including DocVQA \cite{docvqa}, InfographicVQA \cite{infovqa}, DeepForm \cite{svetlichnaya2020deepform} and KLC \cite{klc} for document comprehension, WTQ \cite{wtq} and TabFact \cite{2019TabFactA} for table understanding, ChartQA \cite{chartqa} for chart comprehension, TextVQA \cite{textvqa} and TextCaps \cite{textcaps} for natural image interpretation, VisualMRC \cite{visualmrc} for webpage understanding. For metrics, we use ANLS \cite{anls} for DocVQA and InfoVQA, F1 score for DeepForm and KLC, text-matching accuracy for WTQ, TabFact, TextVQA and ChartQA, CIDEr \cite{cider} for TextCaps and VisualMRC.

As detailed in \cref{tab:maintable}, we observe that DOGR achieves competitive performance across all tasks. Notably, DOGR outperforms the previous state-of-the-art model IXC2.5 by +5.3\% in accuracy on WTQ and +25.0\% in CIDEr on VisualMRC.
It is important to highlight that our primary focus is on enhancing the grounding and referencing capabilities, and we do not engage in extensive pre-training and fine-tuning on large datasets as InternVL2 \cite{chen2024far} and IXC2.5 \cite{internlmxcomposer2_5}.
Despite the limited data used for training, DOGR ranks second on three datasets while remaining highly competitive, showing that DOGR maintains a strong performance across general document understanding tasks.
We believe that incorporating more data for pre-training and fine-tuning could further improve DOGR's performance on these tasks.

\subsection{Ablation Study}
\label{sec:ablation}

\noindent
{\bf Effectiveness of multi-granular parsing data.} 
We evaluate the text recognition and localization performance on DocLocal4K~\cite{mplug1.5}.
We set our model pre-trained with only DocStruct4M\cite{mplug1.5} as our baseline. 
As shown in \cref{tab:stage1_abl2}, with our multi-granular parsing data, the model achieves performance improvements at all granularities compared to the baseline.
Despite the domain gap between our parsing data and DocLocal4K due to the different construction methods, it is evident that incorporating our data improves text grounding and recognition accuracy across various text granularities. 
This validates the effectiveness of our multi-granular parsing data in enhancing basic text localization and recognition capabilities.

\noindent
{\bf Effectiveness of instruction-tuning data.} Besides endowing the model with grounding and referring capabilities, instruction-tuning data further promotes the performance in document understanding. The model fine-tuned using data from DocOwl-1.5\cite{mplug1.5} serves as the baseline. As shown in \cref{tab:baseline}, integrating our instruction-tuning data with DocOwl-1.5's data leads to a 2.87\% increase in  ANLS on InfoVQA and a 21.35\% increase  in CIDEr on VisualMRC.

\vspace{-0.1cm}
\section{Conclusion}
\label{sec:conclusion}

In this paper, we introduce DOGR-Engine, a data construction pipeline for generating high-quality ground-and-refer data for fine-grained document understanding. Additionally, we construct DOGR-Bench as the first comprehensive benchmark for evaluating the document grounding and referring capabilities of MLLMs. Furthermore, leveraging data generated by our engine, we develop DOGR, a pioneering MLLM that integrates text grounding and referring abilities into the dialogue and reasoning process. Our results show that DOGR can achieve versatile document grounding and referring while achieving promising performance on traditional document understanding tasks. 
We hope our work will facilitate practical document AI assistants.
\section{Acknowledgement}
\label{sec:ack}
This work was supported by the Early Career Scheme (No. CityU 21219323) and the General Research Fund (No. CityU 11220324) of the University Grants Committee (UGC), the NSFC Young Scientists Fund (No. 9240127), and the Donation for Research Projects (No. 9229164).


{
    \small
    \bibliographystyle{ieeenat_fullname}
    \bibliography{main}

\begin{thebibliography}{66}
\providecommand{\natexlab}[1]{#1}
\providecommand{\url}[1]{\texttt{#1}}
\expandafter\ifx\csname urlstyle\endcsname\relax
  \providecommand{\doi}[1]{doi: #1}\else
  \providecommand{\doi}{doi: \begingroup \urlstyle{rm}\Url}\fi

\bibitem[Bai et~al.(2025)Bai, Chen, Liu, Wang, Ge, Song, Dang, Wang, Wang, Tang, Zhong, Zhu, Yang, Li, Wan, Wang, Ding, Fu, Xu, Ye, Zhang, Xie, Cheng, Zhang, Yang, Xu, and Lin]{Qwen2.5-VL}
Shuai Bai, Keqin Chen, Xuejing Liu, Jialin Wang, Wenbin Ge, Sibo Song, Kai Dang, Peng Wang, Shijie Wang, Jun Tang, Humen Zhong, Yuanzhi Zhu, Mingkun Yang, Zhaohai Li, Jianqiang Wan, Pengfei Wang, Wei Ding, Zheren Fu, Yiheng Xu, Jiabo Ye, Xi Zhang, Tianbao Xie, Zesen Cheng, Hang Zhang, Zhibo Yang, Haiyang Xu, and Junyang Lin.
\newblock Qwen2.5-vl technical report.
\newblock \emph{arXiv preprint arXiv:2502.13923}, 2025.

\bibitem[Biten et~al.(2019)Biten, Tito, Mafla, Gomez, Rusiñol, Jawahar, Valveny, and Karatzas]{anls}
Ali~Furkan Biten, Rubèn Tito, Andrés Mafla, Lluis Gomez, Marçal Rusiñol, C.V. Jawahar, Ernest Valveny, and Dimosthenis Karatzas.
\newblock Scene text visual question answering.
\newblock In \emph{2019 IEEE/CVF International Conference on Computer Vision (ICCV)}, pages 4290--4300, 2019.

\bibitem[Carter(2024)]{textocr-gpt4v}
Jimmy Carter.
\newblock Textocr-gpt4v.
\newblock \url{https://huggingface.co/datasets/jimmycarter/textocr-gpt4v}, 2024.

\bibitem[Chen et~al.(2023{\natexlab{a}})Chen, Zhang, Zeng, Zhang, Zhu, and Zhao]{chen2023shikra}
Keqin Chen, Zhao Zhang, Weili Zeng, Richong Zhang, Feng Zhu, and Rui Zhao.
\newblock Shikra: Unleashing multimodal llm's referential dialogue magic.
\newblock \emph{arXiv preprint arXiv:2306.15195}, 2023{\natexlab{a}}.

\bibitem[Chen et~al.(2020)Chen, Wang, Chen, Zhang, Wang, Li, Zhou, and Wang]{2019TabFactA}
Wenhu Chen, Hongmin Wang, Jianshu Chen, Yunkai Zhang, Hong Wang, Shiyang Li, Xiyou Zhou, and William~Yang Wang.
\newblock Tabfact : A large-scale dataset for table-based fact verification.
\newblock In \emph{International Conference on Learning Representations (ICLR)}, Addis Ababa, Ethiopia, 2020.

\bibitem[Chen et~al.(2023{\natexlab{b}})Chen, Wu, Wang, Su, Chen, Xing, Zhong, Zhang, Zhu, Lu, Li, Luo, Lu, Qiao, and Dai]{chen2023internvl}
Zhe Chen, Jiannan Wu, Wenhai Wang, Weijie Su, Guo Chen, Sen Xing, Muyan Zhong, Qinglong Zhang, Xizhou Zhu, Lewei Lu, Bin Li, Ping Luo, Tong Lu, Yu Qiao, and Jifeng Dai.
\newblock Internvl: Scaling up vision foundation models and aligning for generic visual-linguistic tasks.
\newblock \emph{arXiv preprint arXiv:2312.14238}, 2023{\natexlab{b}}.

\bibitem[Chen et~al.(2024{\natexlab{a}})Chen, Wang, Cao, Liu, Gao, Cui, Zhu, Ye, Tian, Liu, et~al.]{chen2024expanding}
Zhe Chen, Weiyun Wang, Yue Cao, Yangzhou Liu, Zhangwei Gao, Erfei Cui, Jinguo Zhu, Shenglong Ye, Hao Tian, Zhaoyang Liu, et~al.
\newblock Expanding performance boundaries of open-source multimodal models with model, data, and test-time scaling.
\newblock \emph{arXiv preprint arXiv:2412.05271}, 2024{\natexlab{a}}.

\bibitem[Chen et~al.(2024{\natexlab{b}})Chen, Wang, Tian, Ye, Gao, Cui, Tong, Hu, Luo, Ma, et~al.]{chen2024far}
Zhe Chen, Weiyun Wang, Hao Tian, Shenglong Ye, Zhangwei Gao, Erfei Cui, Wenwen Tong, Kongzhi Hu, Jiapeng Luo, Zheng Ma, et~al.
\newblock How far are we to gpt-4v? closing the gap to commercial multimodal models with open-source suites.
\newblock \emph{arXiv preprint arXiv:2404.16821}, 2024{\natexlab{b}}.

\bibitem[Cheng et~al.(2022)Cheng, Dong, Wang, Jia, Guo, Gao, Han, Lou, and Zhang]{cheng2021hitab}
Zhoujun Cheng, Haoyu Dong, Zhiruo Wang, Ran Jia, Jiaqi Guo, Yan Gao, Shi Han, Jian-Guang Lou, and Dongmei Zhang.
\newblock Hitab: A hierarchical table dataset for question answering and natural language generation.
\newblock In \emph{ACL}, 2022.

\bibitem[Comanici et~al.(2025)Comanici, Bieber, Schaekermann, Pasupat, Sachdeva, Dhillon, Blistein, Ram, Zhang, Rosen, et~al.]{comanici2025gemini}
Gheorghe Comanici, Eric Bieber, Mike Schaekermann, Ice Pasupat, Noveen Sachdeva, Inderjit Dhillon, Marcel Blistein, Ori Ram, Dan Zhang, Evan Rosen, et~al.
\newblock Gemini 2.5: Pushing the frontier with advanced reasoning, multimodality, long context, and next generation agentic capabilities.
\newblock \emph{arXiv preprint arXiv:2507.06261}, 2025.

\bibitem[contributors(2024)]{PyMuPDF}
PyMuPDF contributors.
\newblock Pymupdf: Python bindings for mupdf.
\newblock \url{https://github.com/pymupdf/PyMuPDF}, 2024.

\bibitem[Dong et~al.(2024)Dong, Zhang, Zang, Cao, Wang, Ouyang, Zhang, Duan, Zhang, Li, Yan, Gao, Chen, Zhang, Li, Li, Wang, Chen, He, Zhang, Dai, Qiao, Lin, and Wang]{internlmxcomposer2_4khd}
Xiaoyi Dong, Pan Zhang, Yuhang Zang, Yuhang Cao, Bin Wang, Linke Ouyang, Songyang Zhang, Haodong Duan, Wenwei Zhang, Yining Li, Hang Yan, Yang Gao, Zhe Chen, Xinyue Zhang, Wei Li, Jingwen Li, Wenhai Wang, Kai Chen, Conghui He, Xingcheng Zhang, Jifeng Dai, Yu Qiao, Dahua Lin, and Jiaqi Wang.
\newblock Internlm-xcomposer2-4khd: A pioneering large vision-language model handling resolutions from 336 pixels to 4k hd.
\newblock \emph{arXiv preprint arXiv:2404.06512}, 2024.

\bibitem[Hu et~al.(2024{\natexlab{a}})Hu, Xu, Ye, Yan, Zhang, Zhang, Li, Zhang, Jin, Huang, et~al.]{hu2024docowl}
Anwen Hu, Haiyang Xu, Jiabo Ye, Ming Yan, Liang Zhang, Bo Zhang, Chen Li, Ji Zhang, Qin Jin, Fei Huang, et~al.
\newblock mplug-docowl 1.5: Unified structure learning for ocr-free document understanding.
\newblock \emph{arXiv preprint arXiv:2403.12895}, 2024{\natexlab{a}}.

\bibitem[Hu et~al.(2024{\natexlab{b}})Hu, Xu, Ye, Yan, Zhang, Zhang, Li, Zhang, Jin, Huang, et~al.]{mplug1.5}
Anwen Hu, Haiyang Xu, Jiabo Ye, Ming Yan, Liang Zhang, Bo Zhang, Chen Li, Ji Zhang, Qin Jin, Fei Huang, et~al.
\newblock mplug-docowl 1.5: Unified structure learning for ocr-free document understanding.
\newblock \emph{arXiv preprint arXiv:2403.12895}, 2024{\natexlab{b}}.

\bibitem[Hu et~al.(2024{\natexlab{c}})Hu, Xu, Zhang, Ye, Yan, Zhang, Jin, Huang, and Zhou]{mplug2}
Anwen Hu, Haiyang Xu, Liang Zhang, Jiabo Ye, Ming Yan, Ji Zhang, Qin Jin, Fei Huang, and Jingren Zhou.
\newblock mplug-docowl2: High-resolution compressing for ocr-free multi-page document understanding, 2024{\natexlab{c}}.

\bibitem[Hurst et~al.(2024)Hurst, Lerer, Goucher, Perelman, Ramesh, Clark, Ostrow, Welihinda, Hayes, Radford, et~al.]{gpt4o}
Aaron Hurst, Adam Lerer, Adam~P Goucher, Adam Perelman, Aditya Ramesh, Aidan Clark, AJ Ostrow, Akila Welihinda, Alan Hayes, Alec Radford, et~al.
\newblock Gpt-4o system card.
\newblock \emph{arXiv preprint arXiv:2410.21276}, 2024.

\bibitem[Kafle et~al.(2018)Kafle, Price, Cohen, and Kanan]{kafle2018dvqa}
Kushal Kafle, Brian Price, Scott Cohen, and Christopher Kanan.
\newblock Dvqa: Understanding data visualizations via question answering.
\newblock In \emph{CVPR}, 2018.

\bibitem[Kahou et~al.(2018)Kahou, Michalski, Atkinson, Kadar, Trischler, and Bengio]{kahou2018figureqaannotatedfiguredataset}
Samira~Ebrahimi Kahou, Vincent Michalski, Adam Atkinson, Akos Kadar, Adam Trischler, and Yoshua Bengio.
\newblock Figureqa: An annotated figure dataset for visual reasoning, 2018.

\bibitem[Kembhavi et~al.(2016)Kembhavi, Salvato, Kolve, Seo, Hajishirzi, and Farhadi]{kembhavi2016diagram}
Aniruddha Kembhavi, Mike Salvato, Eric Kolve, Minjoon Seo, Hannaneh Hajishirzi, and Ali Farhadi.
\newblock A diagram is worth a dozen images.
\newblock In \emph{ECCV}, 2016.

\bibitem[Kembhavi et~al.(2017)Kembhavi, Seo, Schwenk, Choi, Farhadi, and Hajishirzi]{kembhavi2017you}
Aniruddha Kembhavi, Minjoon Seo, Dustin Schwenk, Jonghyun Choi, Ali Farhadi, and Hannaneh Hajishirzi.
\newblock Are you smarter than a sixth grader? textbook question answering for multimodal machine comprehension.
\newblock In \emph{Proceedings of the IEEE Conference on Computer Vision and Pattern recognition}, pages 4999--5007, 2017.

\bibitem[Kim et~al.(2022)Kim, Hong, Yim, Nam, Park, Yim, Hwang, Yun, Han, and Park]{kim2022donut}
Geewook Kim, Teakgyu Hong, Moonbin Yim, JeongYeon Nam, Jinyoung Park, Jinyeong Yim, Wonseok Hwang, Sangdoo Yun, Dongyoon Han, and Seunghyun Park.
\newblock Ocr-free document understanding transformer.
\newblock In \emph{European Conference on Computer Vision (ECCV)}, 2022.

\bibitem[Lin et~al.(2025)Lin, Wang, Ge, Ge, Lu, Wei, Zhang, Sun, and Shan]{lin2025toklip}
Haokun Lin, Teng Wang, Yixiao Ge, Yuying Ge, Zhichao Lu, Ying Wei, Qingfu Zhang, Zhenan Sun, and Ying Shan.
\newblock Toklip: Marry visual tokens to clip for multimodal comprehension and generation.
\newblock \emph{arXiv preprint arXiv:2505.05422}, 2025.

\bibitem[Lin et~al.(2024)Lin, Wei, An, Gao, Zou, Luo, Huang, Zhang, and Li]{lin2024drawandunderstand}
Weifeng Lin, Xinyu Wei, Ruichuan An, Peng Gao, Bocheng Zou, Yulin Luo, Siyuan Huang, Shanghang Zhang, and Hongsheng Li.
\newblock Draw-and-understand: Leveraging visual prompts to enable mllms to comprehend what you want, 2024.

\bibitem[Liu et~al.(2024{\natexlab{a}})Liu, Wei, Chen, Kong, Ge, Zhu, Zhao, Sun, Han, and Zhang]{fox}
Chenglong Liu, Haoran Wei, Jinyue Chen, Lingyu Kong, Zheng Ge, Zining Zhu, Liang Zhao, Jianjian Sun, Chunrui Han, and Xiangyu Zhang.
\newblock Focus anywhere for fine-grained multi-page document understanding.
\newblock \emph{arXiv preprint arXiv:2405.14295}, 2024{\natexlab{a}}.

\bibitem[Liu et~al.(2024{\natexlab{b}})Liu, Wei, Chen, Kong, Ge, Zhu, Zhao, Sun, Han, and Zhang]{liu2024focus}
Chenglong Liu, Haoran Wei, Jinyue Chen, Lingyu Kong, Zheng Ge, Zining Zhu, Liang Zhao, Jianjian Sun, Chunrui Han, and Xiangyu Zhang.
\newblock Focus anywhere for fine-grained multi-page document understanding.
\newblock \emph{arXiv preprint arXiv:2405.14295}, 2024{\natexlab{b}}.

\bibitem[Liu et~al.(2023{\natexlab{a}})Liu, Lin, Li, Wang, Yacoob, and Wang]{liu2023aligning}
Fuxiao Liu, Kevin Lin, Linjie Li, Jianfeng Wang, Yaser Yacoob, and Lijuan Wang.
\newblock Aligning large multi-modal model with robust instruction tuning.
\newblock \emph{arXiv preprint arXiv:2306.14565}, 2023{\natexlab{a}}.

\bibitem[Liu et~al.(2023{\natexlab{b}})Liu, Wang, Yao, Chen, Song, Cho, Yacoob, and Yu]{liu2023mmc}
Fuxiao Liu, Xiaoyang Wang, Wenlin Yao, Jianshu Chen, Kaiqiang Song, Sangwoo Cho, Yaser Yacoob, and Dong Yu.
\newblock Mmc: Advancing multimodal chart understanding with large-scale instruction tuning.
\newblock \emph{arXiv preprint arXiv:2311.10774}, 2023{\natexlab{b}}.

\bibitem[Liu et~al.(2023{\natexlab{c}})Liu, Li, Wu, and Lee]{llava}
Haotian Liu, Chunyuan Li, Qingyang Wu, and Yong~Jae Lee.
\newblock Visual instruction tuning, 2023{\natexlab{c}}.

\bibitem[Liu et~al.(2024{\natexlab{c}})Liu, Yang, Liu, Li, Ma, Zhang, and Bai]{liu2024textmonkeyocrfreelargemultimodal}
Yuliang Liu, Biao Yang, Qiang Liu, Zhang Li, Zhiyin Ma, Shuo Zhang, and Xiang Bai.
\newblock Textmonkey: An ocr-free large multimodal model for understanding document, 2024{\natexlab{c}}.

\bibitem[Lv et~al.(2023)Lv, Huang, Chen, Cui, Ma, Chang, Huang, Wang, Dong, Luo, et~al.]{lv2023kosmos}
Tengchao Lv, Yupan Huang, Jingye Chen, Lei Cui, Shuming Ma, Yaoyao Chang, Shaohan Huang, Wenhui Wang, Li Dong, Weiyao Luo, et~al.
\newblock Kosmos-2.5: A multimodal literate model.
\newblock \emph{arXiv preprint arXiv:2309.11419}, 2023.

\bibitem[Marti and Bunke(2002)]{marti2002iam}
U-V Marti and Horst Bunke.
\newblock The iam-database: an english sentence database for offline handwriting recognition.
\newblock \emph{International journal on document analysis and recognition}, 5:\penalty0 39--46, 2002.

\bibitem[Masry et~al.(2022{\natexlab{a}})Masry, Do, Tan, Joty, and Hoque]{chartqa}
Ahmed Masry, Xuan~Long Do, Jia~Qing Tan, Shafiq Joty, and Enamul Hoque.
\newblock {C}hart{QA}: A benchmark for question answering about charts with visual and logical reasoning.
\newblock In \emph{Findings of the Association for Computational Linguistics: ACL 2022}, pages 2263--2279, Dublin, Ireland, 2022{\natexlab{a}}. Association for Computational Linguistics.

\bibitem[Masry et~al.(2022{\natexlab{b}})Masry, Do, Tan, Joty, and Hoque]{masry2022chartqa}
Ahmed Masry, Xuan~Long Do, Jia~Qing Tan, Shafiq Joty, and Enamul Hoque.
\newblock Chartqa: A benchmark for question answering about charts with visual and logical reasoning.
\newblock In \emph{Findings of the Association for Computational Linguistics: ACL 2022}, pages 2263--2279, 2022{\natexlab{b}}.

\bibitem[Mathew et~al.(2021)Mathew, Karatzas, and Jawahar]{docvqa}
Minesh Mathew, Dimosthenis Karatzas, and C.V. Jawahar.
\newblock Docvqa: A dataset for vqa on document images.
\newblock In \emph{Proceedings of the IEEE/CVF Winter Conference on Applications of Computer Vision (WACV)}, pages 2200--2209, 2021.

\bibitem[Mathew et~al.(2022)Mathew, Bagal, Tito, Karatzas, Valveny, and Jawahar]{infovqa}
Minesh Mathew, Viraj Bagal, Rubèn Tito, Dimosthenis Karatzas, Ernest Valveny, and C.~V. Jawahar.
\newblock Infographicvqa.
\newblock In \emph{2022 IEEE/CVF Winter Conference on Applications of Computer Vision (WACV)}, pages 2582--2591, 2022.

\bibitem[Mishra et~al.(2012)Mishra, Alahari, and Jawahar]{MishraBMVC12}
A. Mishra, K. Alahari, and C.~V. Jawahar.
\newblock Scene text recognition using higher order language priors.
\newblock In \emph{BMVC}, 2012.

\bibitem[Obeid and Hoque(2020)]{obeid2020charttotextgeneratingnaturallanguage}
Jason Obeid and Enamul Hoque.
\newblock Chart-to-text: Generating natural language descriptions for charts by adapting the transformer model, 2020.

\bibitem[Pasupat and Liang(2015)]{wtq}
Panupong Pasupat and Percy Liang.
\newblock Compositional semantic parsing on semi-structured tables.
\newblock In \emph{Proceedings of the 53rd Annual Meeting of the Association for Computational Linguistics and the 7th International Joint Conference on Natural Language Processing (Volume 1: Long Papers)}, pages 1470--1480, Beijing, China, 2015. Association for Computational Linguistics.

\bibitem[Peng et~al.(2023)Peng, Wang, Dong, Hao, Huang, Ma, and Wei]{kosmos-2}
Zhiliang Peng, Wenhui Wang, Li Dong, Yaru Hao, Shaohan Huang, Shuming Ma, and Furu Wei.
\newblock Kosmos-2: Grounding multimodal large language models to the world.
\newblock \emph{ArXiv}, abs/2306, 2023.

\bibitem[Rasheed et~al.(2024)Rasheed, Maaz, Shaji, Shaker, Khan, Cholakkal, Anwer, Xing, Yang, and Khan]{hanoona2023GLaMM}
Hanoona Rasheed, Muhammad Maaz, Sahal Shaji, Abdelrahman Shaker, Salman Khan, Hisham Cholakkal, Rao~M. Anwer, Eric Xing, Ming-Hsuan Yang, and Fahad~S. Khan.
\newblock Glamm: Pixel grounding large multimodal model.
\newblock \emph{The IEEE/CVF Conference on Computer Vision and Pattern Recognition}, 2024.

\bibitem[Sidorov et~al.(2020)Sidorov, Hu, Rohrbach, and Singh]{textcaps}
Oleksii Sidorov, Ronghang Hu, Marcus Rohrbach, and Amanpreet Singh.
\newblock Textcaps: a dataset for image captioningwith reading comprehension.
\newblock In \emph{European Conference on Computer Vision}, 2020.

\bibitem[Singh et~al.(2019)Singh, Natarajan, Shah, Jiang, Chen, Batra, Parikh, and Rohrbach]{textvqa}
Amanpreet Singh, Vivek Natarajan, Meet Shah, Yu Jiang, Xinlei Chen, Dhruv Batra, Devi Parikh, and Marcus Rohrbach.
\newblock Towards vqa models that can read.
\newblock In \emph{2019 IEEE/CVF Conference on Computer Vision and Pattern Recognition (CVPR)}, pages 8309--8318, 2019.

\bibitem[Stanis{\l}awek et~al.(2021)Stanis{\l}awek, Grali{\'{n}}ski, Wr{\'o}blewska, Lipi{\'{n}}ski, Kaliska, Rosalska, Topolski, and Biecek]{klc}
Tomasz Stanis{\l}awek, Filip Grali{\'{n}}ski, Anna Wr{\'o}blewska, Dawid Lipi{\'{n}}ski, Agnieszka Kaliska, Paulina Rosalska, Bartosz Topolski, and Przemys{\l}aw Biecek.
\newblock Kleister: Key information extraction datasets involving long documents with complex layouts.
\newblock In \emph{Document Analysis and Recognition -- ICDAR 2021}, pages 564--579, Cham, 2021. Springer International Publishing.

\bibitem[Svetlichnaya(2020)]{svetlichnaya2020deepform}
S Svetlichnaya.
\newblock Deepform: Understand structured documents at scale.
\newblock 2020.

\bibitem[Tanaka et~al.(2021)Tanaka, Nishida, and Yoshida]{visualmrc}
Ryota Tanaka, Kyosuke Nishida, and Sen Yoshida.
\newblock Visualmrc: Machine reading comprehension on document images.
\newblock In \emph{AAAI}, 2021.

\bibitem[Tang et~al.(2023)Tang, Boggust, and Satyanarayan]{tang2023vistextbenchmarksemanticallyrich}
Benny~J. Tang, Angie Boggust, and Arvind Satyanarayan.
\newblock Vistext: A benchmark for semantically rich chart captioning, 2023.

\bibitem[Team et~al.(2024)Team, Georgiev, Lei, Burnell, Bai, Gulati, Tanzer, Vincent, Pan, Wang, et~al.]{gemini}
Gemini Team, Petko Georgiev, Ving~Ian Lei, Ryan Burnell, Libin Bai, Anmol Gulati, Garrett Tanzer, Damien Vincent, Zhufeng Pan, Shibo Wang, et~al.
\newblock Gemini 1.5: Unlocking multimodal understanding across millions of tokens of context.
\newblock \emph{arXiv preprint arXiv:2403.05530}, 2024.

\bibitem[Tim~Allison(2024)]{CC-MAIN-2021-31-PDF-UNTRUNCATED}
Peter~Wyatt Tim~Allison.
\newblock Cc-main-2021-31-pdf-untruncated.
\newblock \url{https://github.com/tballison/CC-MAIN-2021-31-PDF-UNTRUNCATED}, 2024.

\bibitem[Tong et~al.(2024)Tong, Brown, Wu, Woo, Middepogu, Akula, Yang, Yang, Iyer, Pan, et~al.]{tong2024cambrian}
Shengbang Tong, Ellis Brown, Penghao Wu, Sanghyun Woo, Manoj Middepogu, Sai~Charitha Akula, Jihan Yang, Shusheng Yang, Adithya Iyer, Xichen Pan, et~al.
\newblock Cambrian-1: A fully open, vision-centric exploration of multimodal llms.
\newblock \emph{arXiv preprint arXiv:2406.16860}, 2024.

\bibitem[Vedantam et~al.(2015)Vedantam, Zitnick, and Parikh]{cider}
Ramakrishna Vedantam, C.~Lawrence Zitnick, and Devi Parikh.
\newblock Cider: Consensus-based image description evaluation.
\newblock In \emph{2015 IEEE Conference on Computer Vision and Pattern Recognition (CVPR)}, pages 4566--4575, 2015.

\bibitem[Wang et~al.(2021)Wang, Li, Zhou, Chen, Grossman, and Li]{wang2021screen2wordsautomaticmobileui}
Bryan Wang, Gang Li, Xin Zhou, Zhourong Chen, Tovi Grossman, and Yang Li.
\newblock Screen2words: Automatic mobile ui summarization with multimodal learning, 2021.

\bibitem[Wang et~al.(2024{\natexlab{a}})Wang, Xu, Zhao, Ouyang, Wu, Zhao, Xu, Liu, Qu, Shang, Zhang, Wei, Sui, Li, Shi, Qiao, Lin, and He]{wang2024mineruopensourcesolutionprecise}
Bin Wang, Chao Xu, Xiaomeng Zhao, Linke Ouyang, Fan Wu, Zhiyuan Zhao, Rui Xu, Kaiwen Liu, Yuan Qu, Fukai Shang, Bo Zhang, Liqun Wei, Zhihao Sui, Wei Li, Botian Shi, Yu Qiao, Dahua Lin, and Conghui He.
\newblock Mineru: An open-source solution for precise document content extraction, 2024{\natexlab{a}}.

\bibitem[Wang et~al.(2024{\natexlab{b}})Wang, Bai, Tan, Wang, Fan, Bai, Chen, Liu, Wang, Ge, Fan, Dang, Du, Ren, Men, Liu, Zhou, Zhou, and Lin]{Qwen2-VL}
Peng Wang, Shuai Bai, Sinan Tan, Shijie Wang, Zhihao Fan, Jinze Bai, Keqin Chen, Xuejing Liu, Jialin Wang, Wenbin Ge, Yang Fan, Kai Dang, Mengfei Du, Xuancheng Ren, Rui Men, Dayiheng Liu, Chang Zhou, Jingren Zhou, and Junyang Lin.
\newblock Qwen2-vl: Enhancing vision-language model's perception of the world at any resolution.
\newblock \emph{arXiv preprint arXiv:2409.12191}, 2024{\natexlab{b}}.

\bibitem[Wei et~al.(2023)Wei, Kong, Chen, Zhao, Ge, Yang, Sun, Han, and Zhang]{wei2023vary}
Haoran Wei, Lingyu Kong, Jinyue Chen, Liang Zhao, Zheng Ge, Jinrong Yang, Jianjian Sun, Chunrui Han, and Xiangyu Zhang.
\newblock Vary: Scaling up the vision vocabulary for large vision-language models.
\newblock \emph{arXiv preprint arXiv:2312.06109}, 2023.

\bibitem[Wendler(2023)]{RenderedText}
Chris Wendler.
\newblock wendlerc/renderedtext, 2023.

\bibitem[Xu et~al.(2024)Xu, Jiang, Niu, Deng, Poovendran, Choi, and Lin]{Xu2024MagpieAD}
Zhangchen Xu, Fengqing Jiang, Luyao Niu, Yuntian Deng, Radha Poovendran, Yejin Choi, and Bill~Yuchen Lin.
\newblock Magpie: Alignment data synthesis from scratch by prompting aligned llms with nothing.
\newblock \emph{ArXiv}, abs/2406.08464, 2024.

\bibitem[Yamaguchi(2021)]{yamaguchi2021canvasvae}
Kota Yamaguchi.
\newblock Canvasvae: Learning to generate vector graphic documents.
\newblock In \emph{Proceedings of the IEEE/CVF International Conference on Computer Vision}, pages 5481--5489, 2021.

\bibitem[Yang et~al.(2024)Yang, Yang, Hui, Zheng, Yu, Zhou, Li, Li, Liu, Huang, et~al.]{yang2024qwen2}
An Yang, Baosong Yang, Binyuan Hui, Bo Zheng, Bowen Yu, Chang Zhou, Chengpeng Li, Chengyuan Li, Dayiheng Liu, Fei Huang, et~al.
\newblock Qwen2 technical report.
\newblock \emph{arXiv preprint arXiv:2407.10671}, 2024.

\bibitem[Ye et~al.(2023)Ye, Hu, Xu, Ye, Yan, Xu, Li, Tian, Qian, Zhang, et~al.]{ye2023ureader}
Jiabo Ye, Anwen Hu, Haiyang Xu, Qinghao Ye, Ming Yan, Guohai Xu, Chenliang Li, Junfeng Tian, Qi Qian, Ji Zhang, et~al.
\newblock Ureader: Universal ocr-free visually-situated language understanding with multimodal large language model.
\newblock In \emph{Findings of the Association for Computational Linguistics: EMNLP 2023}, pages 2841--2858, 2023.

\bibitem[You et~al.(2023)You, Zhang, Gan, Du, Zhang, Wang, Cao, Chang, and Yang]{you2023ferret}
Haoxuan You, Haotian Zhang, Zhe Gan, Xianzhi Du, Bowen Zhang, Zirui Wang, Liangliang Cao, Shih-Fu Chang, and Yinfei Yang.
\newblock Ferret: Refer and ground anything anywhere at any granularity.
\newblock \emph{arXiv preprint arXiv:2310.07704}, 2023.

\bibitem[Yuan et~al.(2022)Yuan, Liu, Dikubab, Liu, Ji, Wu, and Bai]{yuan2022syntax}
Ye Yuan, Xiao Liu, Wondimu Dikubab, Hui Liu, Zhilong Ji, Zhongqin Wu, and Xiang Bai.
\newblock Syntax-aware network for handwritten mathematical expression recognition.
\newblock \emph{arXiv preprint arXiv:2203.01601}, 2022.

\bibitem[Zhang et~al.(2024{\natexlab{a}})Zhang, Gao, Gan, Dufter, Wenzel, Huang, Shah, Du, Zhang, Li, et~al.]{zhang2024mm1}
Haotian Zhang, Mingfei Gao, Zhe Gan, Philipp Dufter, Nina Wenzel, Forrest Huang, Dhruti Shah, Xianzhi Du, Bowen Zhang, Yanghao Li, et~al.
\newblock Mm1. 5: Methods, analysis \& insights from multimodal llm fine-tuning.
\newblock \emph{arXiv preprint arXiv:2409.20566}, 2024{\natexlab{a}}.

\bibitem[Zhang et~al.(2025)Zhang, Li, Li, Ren, Zou, Liu, Huang, Gao, Li, Yang, et~al.]{zhang2025llava}
Hao Zhang, Hongyang Li, Feng Li, Tianhe Ren, Xueyan Zou, Shilong Liu, Shijia Huang, Jianfeng Gao, Chunyuan Li, Jainwei Yang, et~al.
\newblock Llava-grounding: Grounded visual chat with large multimodal models.
\newblock In \emph{European Conference on Computer Vision}, pages 19--35. Springer, 2025.

\bibitem[Zhang et~al.(2024{\natexlab{b}})Zhang, Dong, Zang, Cao, Qian, Chen, Guo, Duan, Wang, Ouyang, Zhang, Zhang, Li, Gao, Sun, Zhang, Li, Li, Wang, Yan, He, Zhang, Chen, Dai, Qiao, Lin, and Wang]{internlmxcomposer2_5}
Pan Zhang, Xiaoyi Dong, Yuhang Zang, Yuhang Cao, Rui Qian, Lin Chen, Qipeng Guo, Haodong Duan, Bin Wang, Linke Ouyang, Songyang Zhang, Wenwei Zhang, Yining Li, Yang Gao, Peng Sun, Xinyue Zhang, Wei Li, Jingwen Li, Wenhai Wang, Hang Yan, Conghui He, Xingcheng Zhang, Kai Chen, Jifeng Dai, Yu Qiao, Dahua Lin, and Jiaqi Wang.
\newblock Internlm-xcomposer-2.5: A versatile large vision language model supporting long-contextual input and output.
\newblock \emph{arXiv preprint arXiv:2407.03320}, 2024{\natexlab{b}}.

\bibitem[Zhao et~al.(2023)Zhao, Zhao, Nan, Qi, Zhang, Tang, Mi, and Radev]{zhao-etal-2023-robut}
Yilun Zhao, Chen Zhao, Linyong Nan, Zhenting Qi, Wenlin Zhang, Xiangru Tang, Boyu Mi, and Dragomir Radev.
\newblock {R}obu{T}: A systematic study of table {QA} robustness against human-annotated adversarial perturbations.
\newblock In \emph{Proceedings of the 61st Annual Meeting of the Association for Computational Linguistics (Volume 1: Long Papers)}, pages 6064--6081, Toronto, Canada, 2023. Association for Computational Linguistics.

\bibitem[Zhou et~al.(2025)Zhou, Wang, Lin, Ma, Zhu, and Zheng]{zhou2025scale}
Yinan Zhou, Yaxiong Wang, Haokun Lin, Chen Ma, Li Zhu, and Zhedong Zheng.
\newblock Scale up composed image retrieval learning via modification text generation.
\newblock \emph{arXiv preprint arXiv:2504.05316}, 2025.

\end{thebibliography}
}


\clearpage
\appendix


\setcounter{page}{1}

\section{The Annotation Results}
\label{sec:annotationresults}

\subsection{Poster and Chart Annotations}
\begin{figure*}[t]
  \centering
  \includegraphics[width=0.9\linewidth]{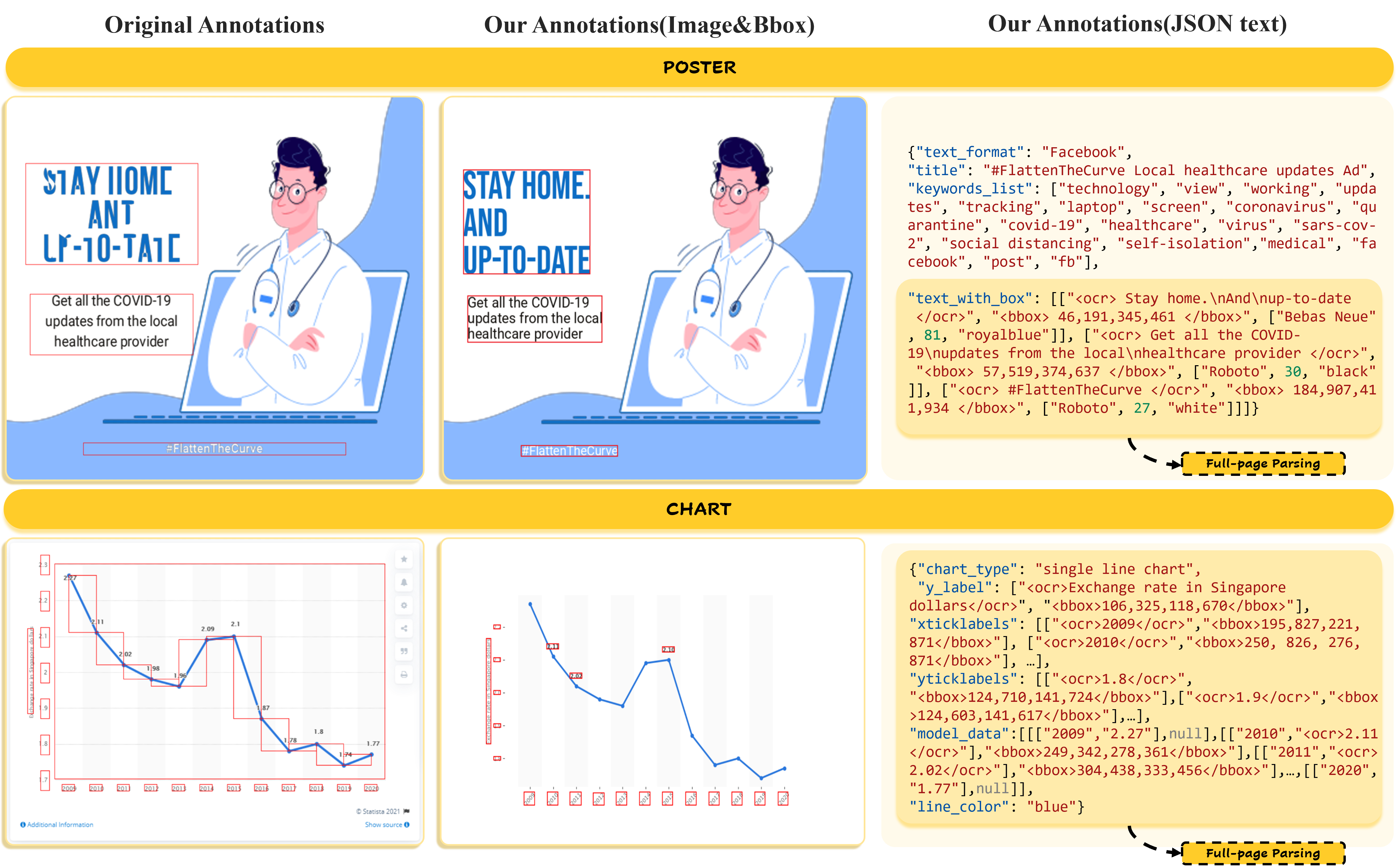}
   \caption{Comparison of origin annotation and our new constructed annotation.}
   \label{fig:annotation1}
\end{figure*}

As shown in \cref{fig:annotation1}, we present a comparison between our annotations and the original annotations. 

In the original poster annotations, some text image layers are damaged, and some bounding boxes are inaccurate, which hinders precise text recognition and localization. In the annotations obtained using our Re-rendering Strategy, we reconstruct the text layers and achieve accurate bounding box annotations. Additionally, we include global content information from the original dataset, such as text format, title and keywords to enhance global awareness during instruction-tuning data generation. We use the values of ``text\_with\_box'' as the annotations for poster's full-page parsing data.

In original chart annotations from ChartQA, some bounding boxes are associated with bars/lines rather than text value. This is inconsistent with our text and bounding box correspondence objectives. Additionally, the bounding boxes in the original annotations are not accurate. In our re-rendered annotations, we align the bounding boxes with the text value, and the bounding boxes are accurate. Furthermore, we randomly erase some of the values to ensure that the model can infer the missing values based on other visual information. We use the entire chart JSON dict as the annotations for the full-page parsing of the chart. Due to space constraints, we omit some of the content using `...'.

\subsection{PDF Document Annotations}
\begin{figure*}[t]
  \centering
  \includegraphics[width=0.9\linewidth]{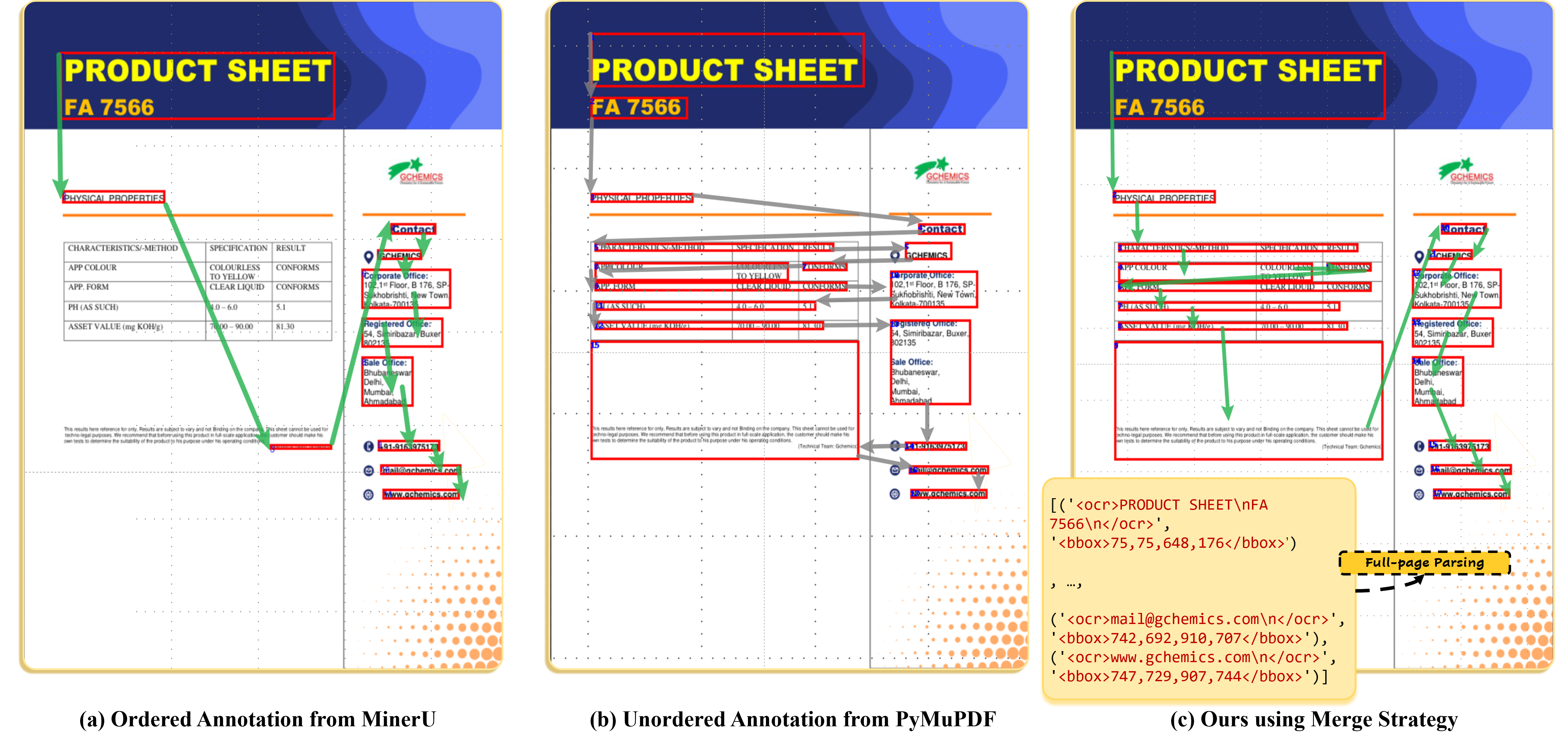}
   \caption{PDF Document Parsing results comparison of ordered annotation from MinerU, unordered annotation from PyMuPDF, and our annotation using the Merge Strategy. \textbf{\textcolor{mygreen2}{Green}} arrow indicates the ordered annotations and \textbf{\textcolor{mygray2}{Gray}} arrow indicates the naive scanning order.}
   \label{fig:annotation2}
\end{figure*}

As shown in \cref{fig:annotation2}, we present a comparison between the ordered annotations from MinerU and the unordered but comprehensive annotations from PyMuPDF, along with the combined annotations using our Merge Strategy. We utilize green arrows to indicate the ordered annotations and gray arrows to indicate the naive scanning order. By combining the two annotation methods, we achieve full-page parsing annotations that are both comprehensive and as ordered as possible. Moreover, our method can become more effective as the performance of the ordered annotation tools improves. Due to space constraints, we omit the content in the middle of this passage.

\section{Prompts and Instructions}
\label{sec:prompts}
\subsection{Prompt Details}

In \cref{fig:prompts}, we show three different prompts for poster, chart and PDF document: 

\begin{figure*}[]
  \centering
  \includegraphics[width=1.0\linewidth]{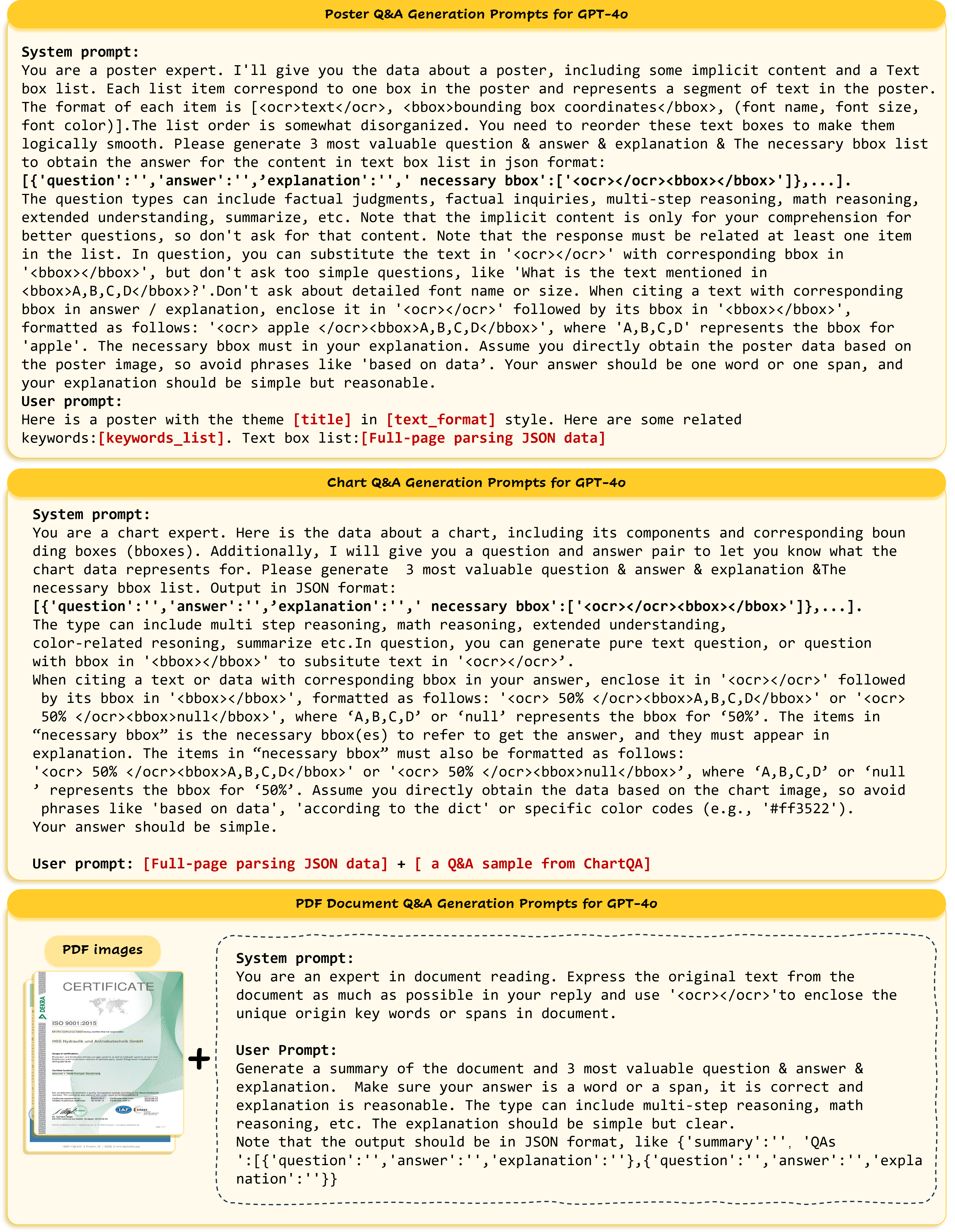}
   \caption{Prompts utilized as input to gpt-4o for poster, chart and PDF document.}
   \label{fig:prompts}

\end{figure*}

For poster, we deploy plain text input as prompt. Besides format information and rule information, we also provide GPT-4o with some of the overall content and style information from the original Crello dataset. This helps in achieving a better global understanding, thereby improving the generation of instruction-tuning data.

For chart, since the content of the charts only contains some numbers and lacks an introduction to the meaning of the content being statistically represented, we add a question answering data from the original ChartQA to help GPT-4o better understand the meaning conveyed by the chart content. Additionally, when the model generates output, we output the masked values in the grounded format ``\texttt{<ocr>text</ocr><bbox>null</bbox>}'' as well. After obtaining the output, we perform format filtering to degrade this part of the content into plain text and remove the degraded plain text item in ``necessary bbox''.

For PDF document, we directly send the images and simple output format rules to GPT-4o, obtaining the output with the original text wrapped in ``\texttt{<ocr></ocr>}''. Then, we use PyMuPDF to query these contents, find the corresponding coordinates, and normalize them. After that, we wrap the coordinates in ``\texttt{<bbox></bbox>}'' and append them to the original wrapped text.

After obtaining the output from GPT-4o, we perform format filtering to remove samples that do not meet the format requirements. And we also perform grounded data checking to correct or remove sampls that have incorrect grounding content. We change the grounded blocks ``\texttt{<ocr></ocr><bbox></bbox>}'' in the questions to ``\texttt{<bbox></bbox>}''. Finally, we combine various answers and reasoning to obtain different types of tasks introduced in \cref{sec:bench}. 

\subsection{Instruction Details}
\begin{figure*}[t]
  \centering
  \includegraphics[width=1.0\linewidth]{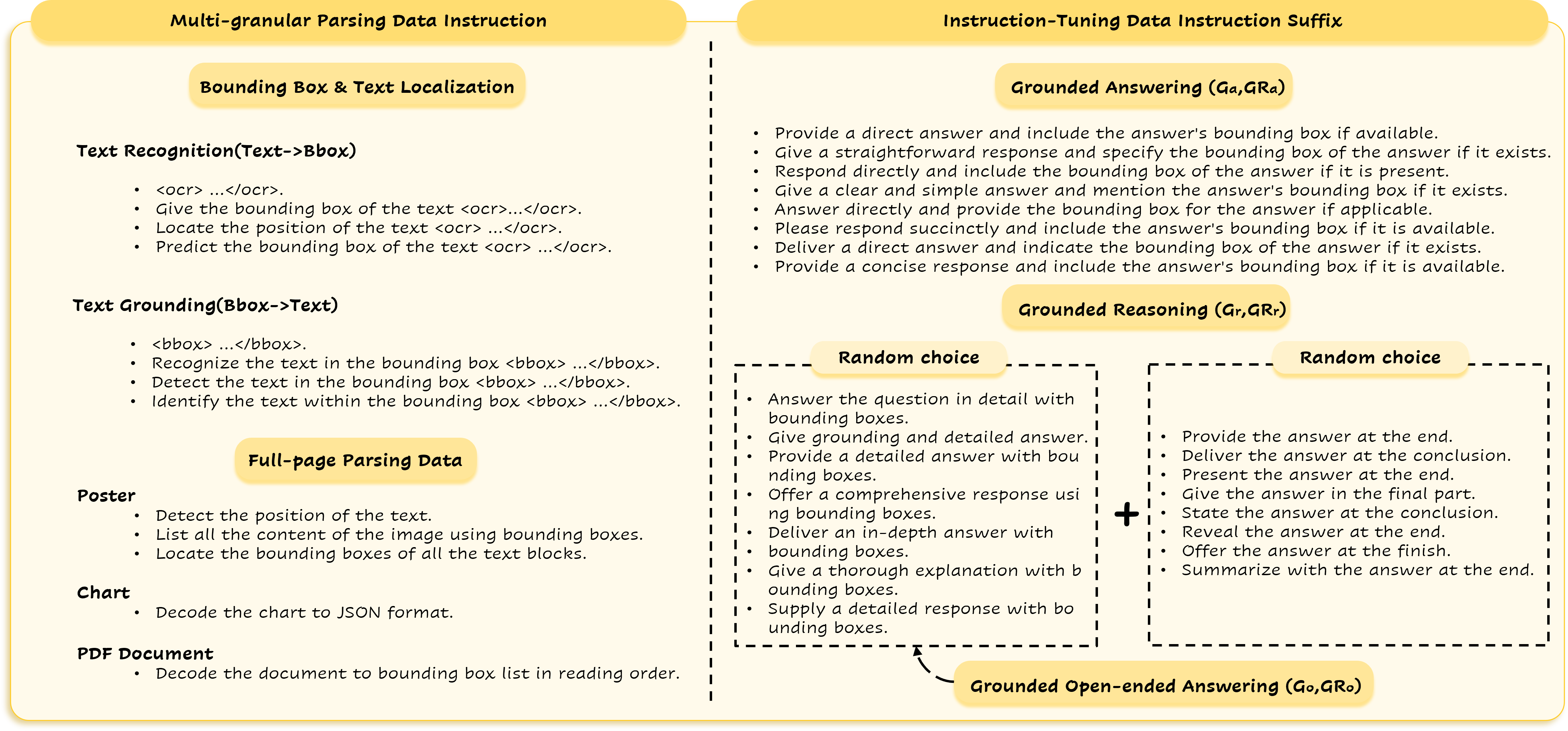}
   \caption{The instruction utilized in Multi-granular Parsing tasks and the response format prompts for instruction-tuning data.}
   \label{fig:suffix}
\end{figure*}

As shown in \cref{fig:suffix}, we introduce the instruction utilized in Multi-granular Parsing tasks and the response format prompts which are followed by the questions for instruction-tuning data. For Bounding Box \& Text Localization, we introduce 4 instructions for each task, the answer is directly wrapped with ``\texttt{<ocr></ocr>}'' or ``\texttt{<bbox></bbox>}''. For Full-page Parsing, we introduce 3 instructions for poster data, 1 for chart, and 1 for PDF document data. The responses are in the format shown in \cref{fig:annotation1} and \cref{fig:annotation2}.

We add different response format prompts to different questions based on the format of the responses to help the model output corresponding results when users interact with it. 
For generated question and answer pairs, we combine different question and answer pairs with various response format prompts to obtain diverse grounded data. For Grounded Answering Data, we have 7 response format prompts to be added after the question, and the answer should be directly a simple text wrapped with ``\texttt{<ocr></ocr>}'' and followed by the coordinates wrapped with ``\texttt{<bbox></bbox>}''. For Grounded Reasoning Data, we combine two random-chose prompts together to form the response format prompt, and the response should be such a sentence structure involves a segment of grounded reasoning followed by "Answer: " and a concise answer. For Grounded Open-ended Answering, we simply use the first part of the response format prompts for Grounded Reasoning, and the response should be a grounded reasoning sentence. For Plain text Answering, we add no response format prompt to keep consistent with the original question answering.

\section{Qualitative Results}
\label{sec:qualitative}

\subsection{Analysis about \cref{fig:teaser}}
We present the inference results of DOGR for grounding, grounding-and-referring, and referring tasks. The specific coordinates of the annotations are omitted, and the grounded text is highlighted with colored boxes. The colors of the boxes within the document image correspond to the colors of the text boxes. For the grounding task, we present four examples. The first sample demonstrates that DOGR can perform fine-grained question answering on general document images with rich content and complex layouts, successfully grounding the corresponding information. The second sample illustrates the model's excellent recognition and localization capabilities for diverse and small text in poster-type images. Additionally, we showcase two samples of grounding in chart-type images, which indicate that DOGR possesses a certain level of mathematical ability, enabling it to provide grounded reasoning during calculation, as well as the capability to estimate values in charts that lack textual annotations based on the axes. For the grounding-and-referring and referring tasks, DOGR is able to recognize the content of user-selected regions and provide reasonable grounded or plain textual reasoning and responses. DOGR exhibits robust fine-grained grounding and referring capabilities, allowing for reliable grounded reasoning and accommodating diverse user interactions, significantly enhancing the overall user experience.

\subsection{More Qualitative Results}

\subsubsection{DOGR-Bench Examples}
\begin{figure*}[t]
  \centering
  \includegraphics[width=1.0\linewidth]{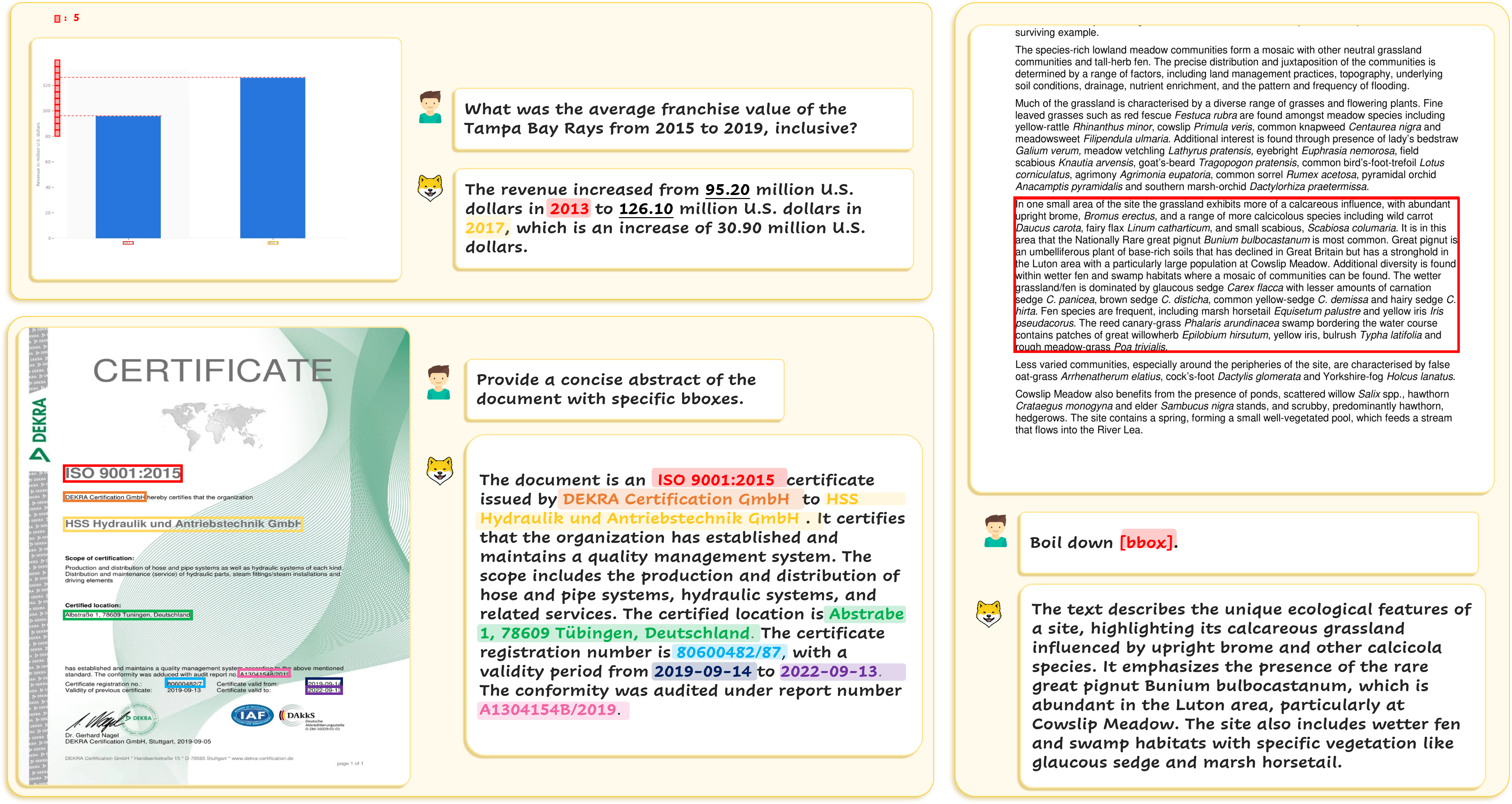}
   \caption{DOGR's inference results on DOGR-Bench.}
   \label{fig:sample_indomain}
   \vspace{-0.5cm}
\end{figure*}

As shown in \cref{fig:sample_indomain}, we present three inference examples of DOGR-Bench, demonstrating unlabeled value reading capability, grounding summarization capability, and referring summarization capability. The first figure is an example of reading unlabeled values. We add auxiliary mark boxes in red to the coordinate axes to facilitate reading the values, each box's height represents 5. It can be observed that the model's output values are accurate. The figure on the lower left is an example of grounding summarization for an entire document. DOGR is able to perceive the content of the entire document and perform grounding summarization effectively. The figure on the right is an example of referring summarization for a specific area. The model can also accomplish this task, thereby helping users improve reading efficiency.

\subsubsection{Other Examples}
\begin{figure*}[t]
  \centering
  \includegraphics[width=1\linewidth]{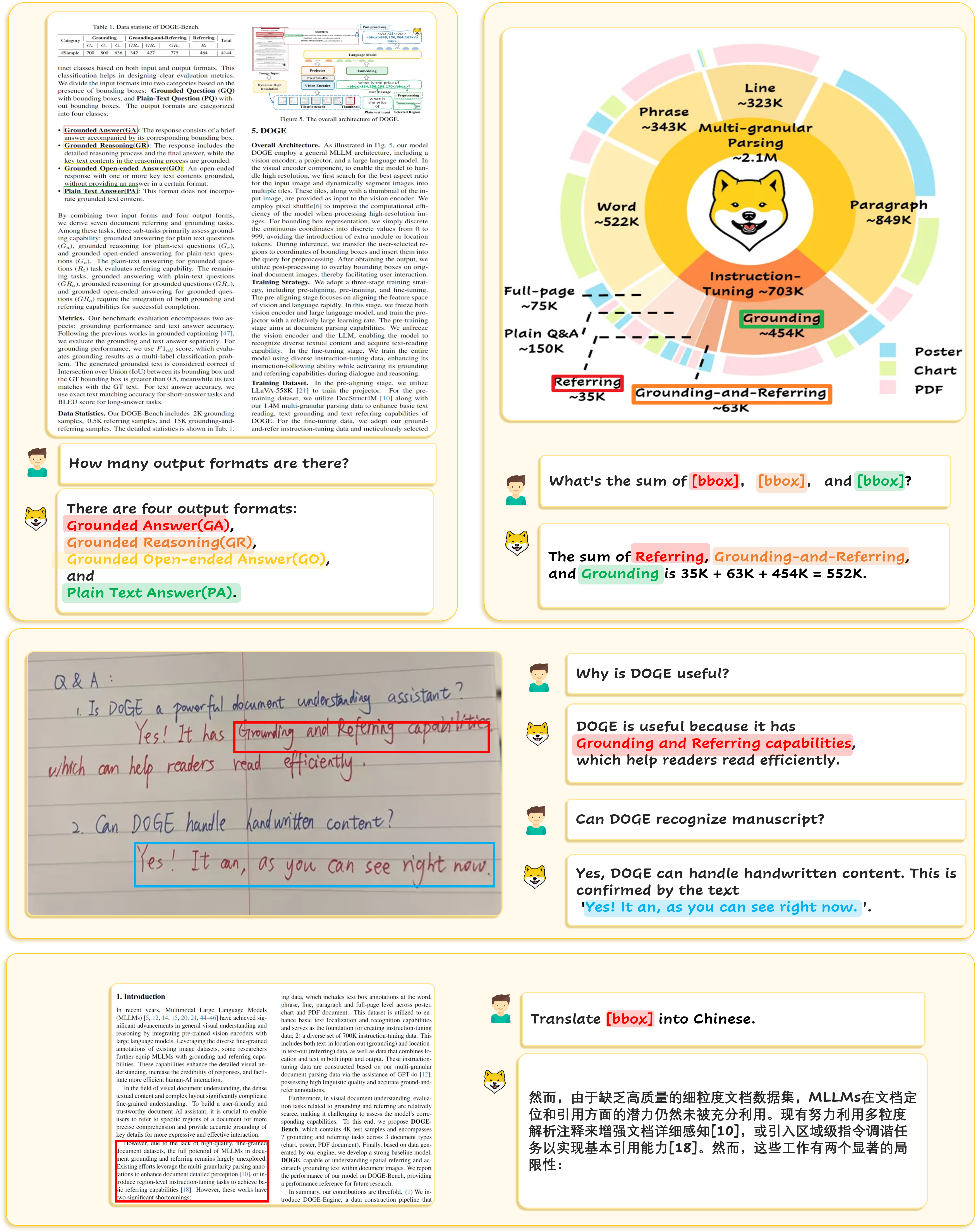}
   \caption{Other DOGR's inference samples.}
   \label{fig:sample_outdomain}
\end{figure*}

It is noteworthy that our model demonstrates strong generalization capabilities beyond some training domains and tasks, as illustrated in \cref{fig:sample_outdomain}.

\noindent
\textbf{Strong generalization.} As shown in the first row of examples, we feed DOGR with screenshots of the paper content and the specially shaped fan chart that the model does not process before. DOGR is able to correctly respond, demonstrating its strong generalization and usability in actual document reading scenarios.

\noindent
\textbf{Handwriting ground-and-refer ability.} The middle sample showcases the model's ability to recognize and grounding handwritten content. DOGR can fully understand this casual handwriting and provide grounded output. It is worth mentioning that our training data does not contain the grounding or referring tasks on such handwritten images.

\noindent
\textbf{Other capabilities.} The bottom sample demonstrates an additional untrained capability of our model, specifically referring translation. DOGR is able to provide translations for the asked region. It is important to note that our training data do not include tasks similar to referring translation, even rarely includes languages other than English. Therefore, we believe that DOGR can effectively handle the relationship between regions and corresponding text, and seamlessly integrate the capabilities of large models with grounding and referring abilities.

\subsubsection{Failure Cases}
\begin{figure*}[t]
  \centering
  \includegraphics[width=1\linewidth]{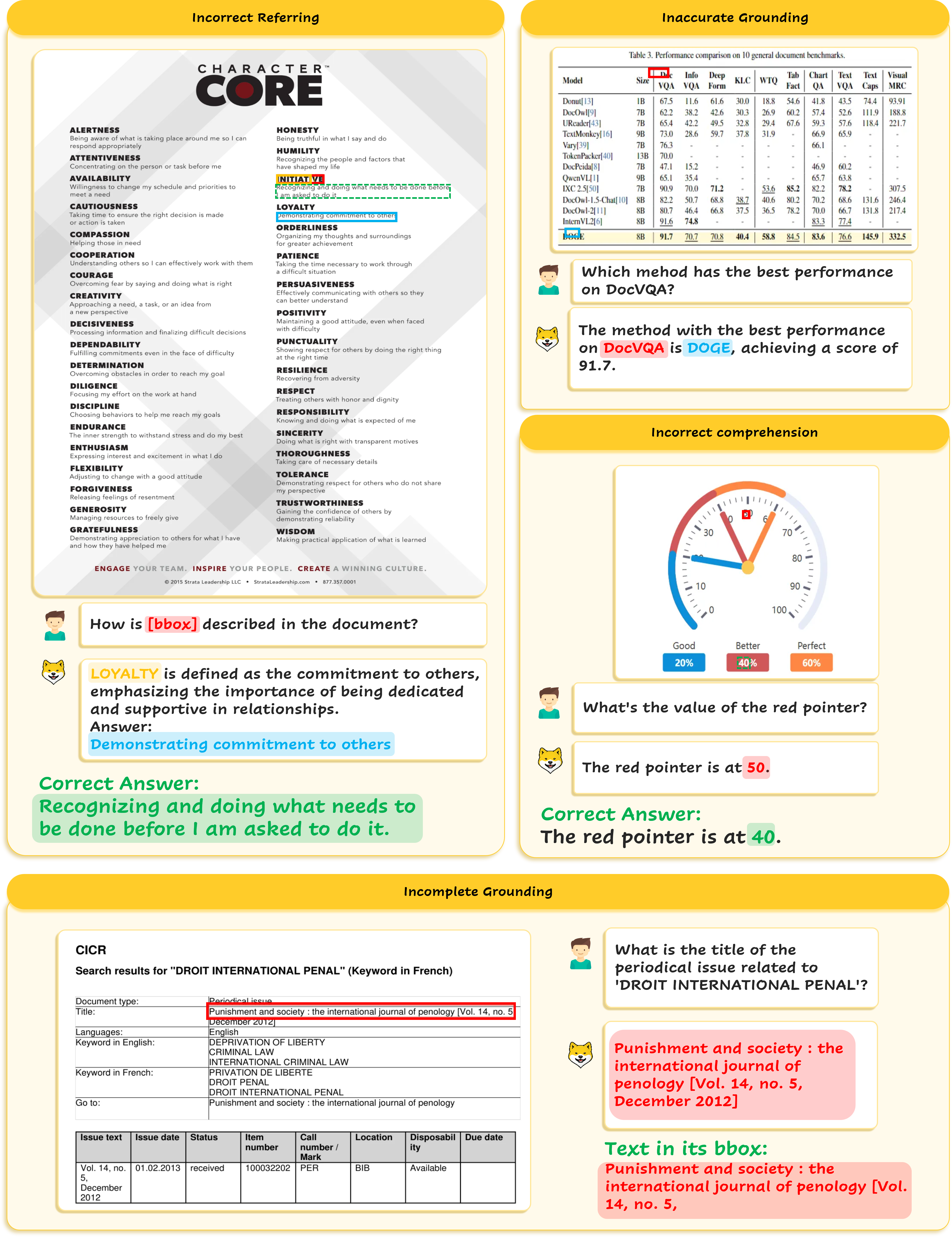}
   \caption{Failure cases of DOGR.}
   \label{fig:sample_failure}
\end{figure*}

Although DOGR demonstrates strong grounding and referring capabilities, there are still some shortcomings, as shown in \cref{fig:sample_failure}.

\noindent
\textbf{Referring.} The upper left figure illustrates an example of incorrect referring. It mistakenly associates the bounding box that should correspond to the text ``INITIATIVE'' in the question with the text ``LOYALTY'' below it. Additionally, its width is re-estimated according to the size corresponding to "LOYALTY," resulting in an incorrect answer.

\noindent
\textbf{Grounding.} When encountering some unfamiliar text content such as tables in the upper right image, DOGR can understand the content, effectively identified the line breaks between ``Doc'' and ``VQA'' and merged them together, and provide correct answers, but the grounding boxes are inaccurate.
There is also a issue of incomplete grounding content, such as the bottom sample, which often occurs when the content requiring grounding is interrupted or wrapped to the next line. Although this does not affect understanding, the text and bounding boxes provided by the model do not completely match. 

\noindent
\textbf{Comprehension.} When facing with some unfamiliar structural document, such as the chart shown in the middle right, the model gives incorrect answers. However, due to the intuitive expressiveness of grounding and referring, readers can quickly determine that the model's answer is incorrect. These samples can also provide evidence of the model's deficiencies, laying the foundation for further improvements.

\clearpage

\section{Dataset Details}
\label{sec:Datasetdetails}

\subsection{Dataset Statistic}
As shown in \cref{tab:count}, we present the detailed composition of the DOGR dataset's Multi-granular Parsing data and Instruction Tuning data. The Multi-granular Parsing data includes five granularities: word, phrase, line, paragraph, and full-page parsing. The Instruction-Tuning data comprises three types of grounded data: grounding referring, grounding-and-referring, and plain text Q\&A. We provide the detailed data volume for each type.

\begin{figure}[]
  \centering
  \includegraphics[width=1\linewidth]{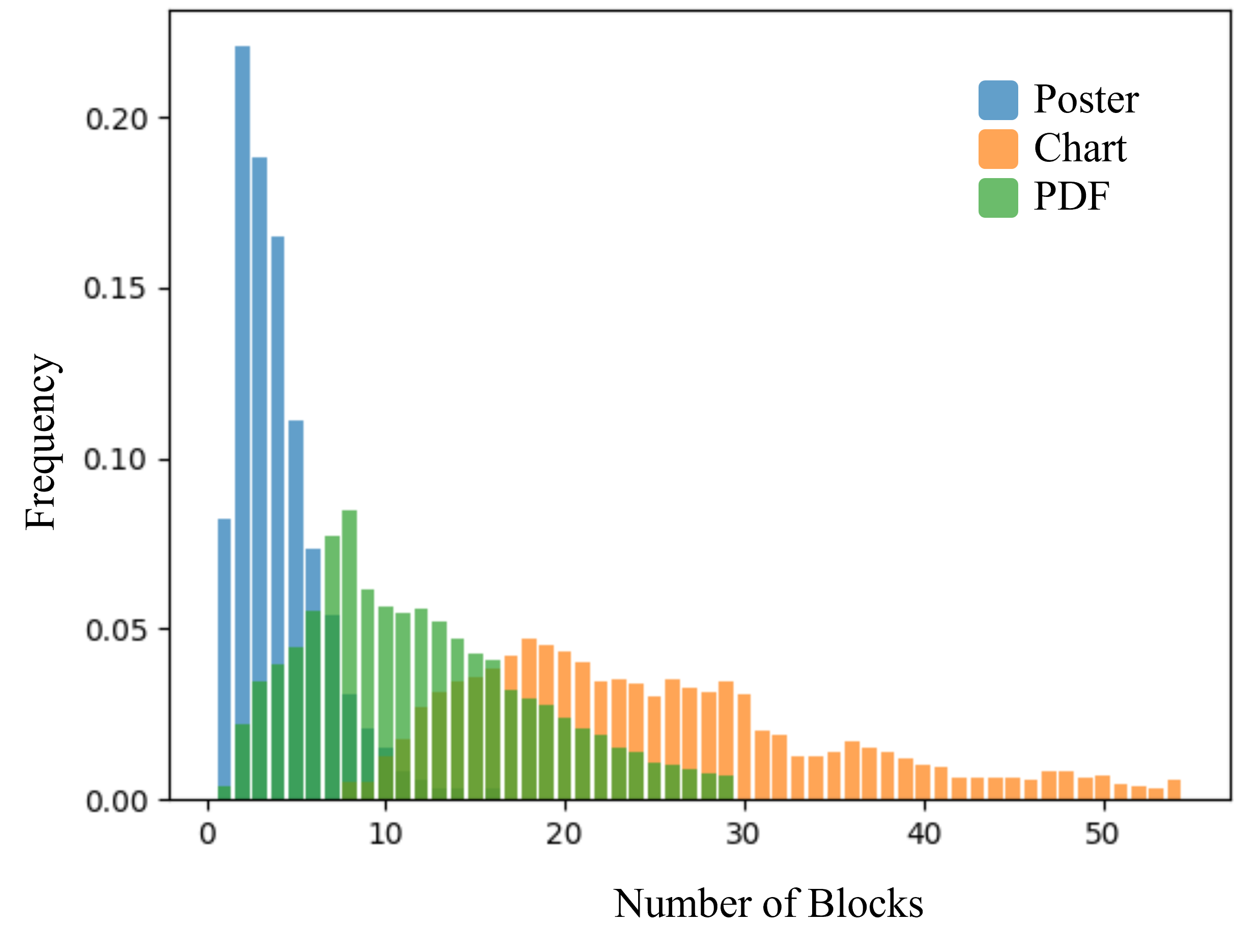}
   \caption{Block count distribution of poster, chart, and PDF.}
   \label{fig:count}

\end{figure}

\noindent
\cref{fig:count} shows the distribution of block counts for poster, chart, and PDF document. Poster has fewer grounded blocks within each image, larger areas, diverse font styles, and larger font sizes. The average text length per block for poster is 3.25 words. Chart, on the other hand, has a higher number of blocks with small areas and small font sizes. Each block in a chart corresponds to a small component value within the chart, with an average text length of 1.34 words per block. PDF document has a moderate distribution of block counts, larger areas, and small font sizes. Each block in a PDF contains relatively longer text, with an average length of 22.55 words per block.

\begin{table}[]
\caption{The detailed composition of the DOGR dataset's Multi-granular parsing data and Instruction-Tuning data.}
\label{tab:count} 
\resizebox{\linewidth}{!}{
\begin{tabular}{c|l|c|c|lc}
\toprule
\textbf{Type}                                                                                        & \multicolumn{1}{c|}{\textbf{Number}} & \textbf{SubType}                         & \textbf{Number}         & \textbf{SubSubType} & \textbf{Number} \\ \midrule
\multirow{15}{*}{\textbf{\begin{tabular}[c]{@{}c@{}}Multi-granular \\ Parsing\end{tabular}}}         & \multirow{15}{*}{2,114,414}            & \multirow{3}{*}{Full Page}               & \multirow{3}{*}{75,391}  & Poster              & 20,867           \\
                                                                                                     &                                      &                                          &                         & Chart               & 31,716           \\
                                                                                                     &                                      &                                          &                         & PDF                 & 22,808           \\ \cmidrule{3-6} 
                                                                                                     &                                      & \multirow{3}{*}{Word}                    & \multirow{3}{*}{522,682} & Poster              & 65,446           \\
                                                                                                     &                                      &                                          &                         & Chart               & 354,731          \\
                                                                                                     &                                      &                                          &                         & PDF                 & 102,505          \\ \cmidrule{3-6} 
                                                                                                     &                                      & \multirow{3}{*}{Span}                    & \multirow{3}{*}{343,596} & Poster              & 56,006           \\
                                                                                                     &                                      &                                          &                         & Chart               & 58,212           \\
                                                                                                     &                                      &                                          &                         & PDF                 & 229,378          \\ \cmidrule{3-6} 
                                                                                                     &                                      & \multirow{3}{*}{Line}                    & \multirow{3}{*}{323,211} & Poster              & 5,577            \\
                                                                                                     &                                      &                                          &                         & Chart               & 6,268            \\
                                                                                                     &                                      &                                          &                         & PDF                 & 311,366          \\ \cmidrule{3-6} 
                                                                                                     &                                      & \multirow{3}{*}{Paragraph}               & \multirow{3}{*}{849,534} & Poster              & 511,998          \\
                                                                                                     &                                      &                                          &                         & Chart               & 229,378          \\
                                                                                                     &                                      &                                          &                         & PDF                 & 108,158          \\ \midrule
\multirow{12}{*}{\textbf{\begin{tabular}[c]{@{}c@{}}Instruction-Tuning\end{tabular}}} & \multirow{12}{*}{703,724}             & \multirow{3}{*}{Grounding}               & \multirow{3}{*}{454,404} & Poster              & 96,718           \\
                                                                                                     &                                      &                                          &                         & Chart               & 62,636           \\
                                                                                                     &                                      &                                          &                         & PDF                 & 295,050          \\ \cmidrule{3-6} 
                                                                                                     &                                      & \multirow{3}{*}{Grounding +   Referring} & \multirow{3}{*}{63,243}  & Poster              & 36,663           \\
                                                                                                     &                                      &                                          &                         & Chart               & 849             \\
                                                                                                     &                                      &                                          &                         & PDF                 & 25,731           \\ \cmidrule{3-6} 
                                                                                                     &                                      & \multirow{3}{*}{Referring}               & \multirow{3}{*}{35,197}  & Poster              & 19,207           \\
                                                                                                     &                                      &                                          &                         & Chart               & 618             \\
                                                                                                     &                                      &                                          &                         & PDF                 & 15,372           \\ \cmidrule{3-6} 
                                                                                                     &                                      & \multirow{3}{*}{Plain Text Q\&A}              & \multirow{3}{*}{150,880} & Poster              & 35,831           \\
                                                                                                     &                                      &                                          &                         & Chart               & 45,948           \\
                                                                                                     &                                      &                                          &                         & PDF                 & 69,101           \\ 
                                                                                                     \bottomrule
\end{tabular}
}
\vspace{-0.5em}
\end{table}

\subsection{Construction Cost Details}
In terms of multi-granular parsing data construction, DOGR-Engine can achieve boundary box annotation and content extraction on any document dataset with a similar rendering process. It can also scale up to a larger volume of PDF source files without additional manual costs. When constructing instruction fine-tuning data, our main cost lies in the API calls to GPT-4o. We use the version gpt-4o-2024-08-06. For poster and chart, our input is long text containing full page parsing data, while for PDF, the input is document images plus short prompt texts. During construction, we can generate multiple question and answer pairs for a single image simultaneously and use batch API requests to save costs. Ultimately, we can construct over 1,000 grounded question and answer pairs for an average cost of \$1, which is far more efficient and cost-effective than manual construction.

\subsection{Hallucinations Avoidance} 
Regarding hallucinations, GPT-4o is quite capable of handling our tasks. By manual checking, we observe that hallucinations occur \textbf{very rarely}. We also filter out samples where the text and bounding boxes do not correspond correctly. Additionally, for DOGR-Bench, we manually filter out the samples that contain hallucinations or other errors.

\section{Training Data Details}
\label{sec:Trainingdetails}

For the pre-training data, we utilize DocStruct4M~\cite{mplug1.5} along with our 2.1M multi-granular parsing data to enhance DOGR's foundational grounding and referring capabilities.

For the fine-tuning data, we compose our 703k Instruction-Tuning data with 575k DocDownStream-1.0 data from \cite{mplug1.5} and 716k other document-related data from various datasets, resulting in a final fine-tuning dataset of 2M. In \cref{tab:dataset_statistics}, we show the detailed data source and sampled number of 716k other document-related data.

\begin{table}[]
\caption{The detailed statistics of 716k other document-related data.}
\label{tab:dataset_statistics}
\centering
\resizebox{\columnwidth}{!}{
    \begin{tabular}{lr|lr}
    \toprule
       
        Dataset & \# Samples &Dataset & \# Samples \\ \midrule
        
        IIIT5K~\cite{MishraBMVC12}  & 1,990 & RoBUT WTQ\cite{zhao-etal-2023-robut}  & 38,241\\
        TextOCR-GPT4V~\cite{textocr-gpt4v} & 25,104 & AI2D (InternVL~\cite{chen2023internvl})  & 12,403 \\
        FigureQA~\cite{kahou2018figureqaannotatedfiguredataset}  & 1,000 &Infographic VQA~\cite{infovqa}  & 8,489 \\
        Diagram Image2Text  & 295 & LRV Chart~\cite{liu2023aligning} & 1,776\\
        K12 Printing  & 20,000&SROIE& 33,616\\
        AI2D (GPT4V Detailed Caption)~\cite{kembhavi2016diagram}  & 4,864& MultiHiertt &7,614\\
        VisText~\cite{tang2023vistextbenchmarksemanticallyrich}  & 9,964&RoBUT WikiSQL  & 74,984\\
        ChartQA~\cite{chartqa}   & 18,260&VisualMRC\cite{visualmrc}  & 3,022\\
        DVQA~\cite{kafle2018dvqa}  & 20,000 &TextCaps~\cite{textcaps}  & 21,942\\
        Magpie Pro~\cite{Xu2024MagpieAD} (L3 ST)  & 50,000&Chart2Text~\cite{obeid2020charttotextgeneratingnaturallanguage}  & 26,956\\
        HiTab~\cite{cheng2021hitab}  & 2,495&HME100K~\cite{yuan2022syntax}  & 74,492\\
        RoBUT SQA  & 8,509&Magpie Pro (L3 MT)&50,000\\
        ChromeWriting~\cite{RenderedText}  & 8,825&Magpie Pro (Qwen2 ST)  & 50,000\\ 
        Screen2Words~\cite{wang2021screen2wordsautomaticmobileui}  & 15,725 &Rendered Text~\cite{RenderedText}  & 9,995\\
        IAM~\cite{marti2002iam}  & 5,658& TQA~\cite{kembhavi2017you}  & 27,302\\
        AI2D (Original)  & 2,429&SynthDog-EN~\cite{kim2022donut}  & 40,000\\
        MMC bInstruction Arxiv\cite{liu2023mmc}&20,000& MMC bInstruction NON-Arxiv &20,000\\
        \bottomrule
    \end{tabular}
    }
    
\end{table}

\begin{table*}[!t]
\centering
\caption{Grounding and recognition performance comparison of the same model pretrained w. and w.o. our pretraining data.}
\label{tab:pretrain_more}
\resizebox{1\linewidth}{!}{

    \begin{tabular}{l|ccccc|ccccc|c}
    \toprule
         & 

    \multicolumn{5}{c|}{\textbf{Poster}} &
    \multicolumn{5}{c|}{\textbf{PDF Document}}&
    \multicolumn{1}{c}{\textbf{Chart}} \\ \midrule
    \multicolumn{12}{c}
    {\textbf{Text Grounding}}      \\ \midrule

    {\textbf{Model}} & word&phrase&line&paragraph&ALL& word&phrase&line&paragraph&ALL&ALL\\\midrule
    DOGR$^{-}$&58.0&63.12&58.96&76.43&63.19&33.88&47.62&51.5&62.88&48.97&\cellcolor{mygray}{24.03}\\ \midrule
    \rowcolor{yellow!10}
    DOGR&\textbf{91.88}&\textbf{94.62}&\textbf{96.28}&\textbf{91.04}&\textbf{93.53}&\textbf{70.25}&\textbf{82.25}&\textbf{87.5}&\textbf{83.62}&\textbf{80.91}&\textbf{86.12}\\ 
    \midrule
    \multicolumn{12}{c}{\textbf{Text Recognition}}      \\\midrule
    DOGR$^{-}$ &83.34&60.96&41.67&36.95&55.73&49.66&58.62&50.34&47.27&51.47&68.86\\\midrule
    \rowcolor{yellow!10}DOGR &\textbf{92.94}&\textbf{94.05}&\textbf{86.52}&\textbf{86.08}&\textbf{89.9}&\textbf{73.57}&\textbf{86.48}&\textbf{76.85}&\textbf{71.8}&\textbf{77.17}&\textbf{95.24}\\
    \bottomrule
    \end{tabular}

}
\end{table*}

\section{Detailed Evaluation of MLLMs on DOGR-Bench.}
\label{sec:Experiments}

\subsection{Evaluation Prompt}
\begin{figure*}[]
  \centering
  \includegraphics[width=1\linewidth]{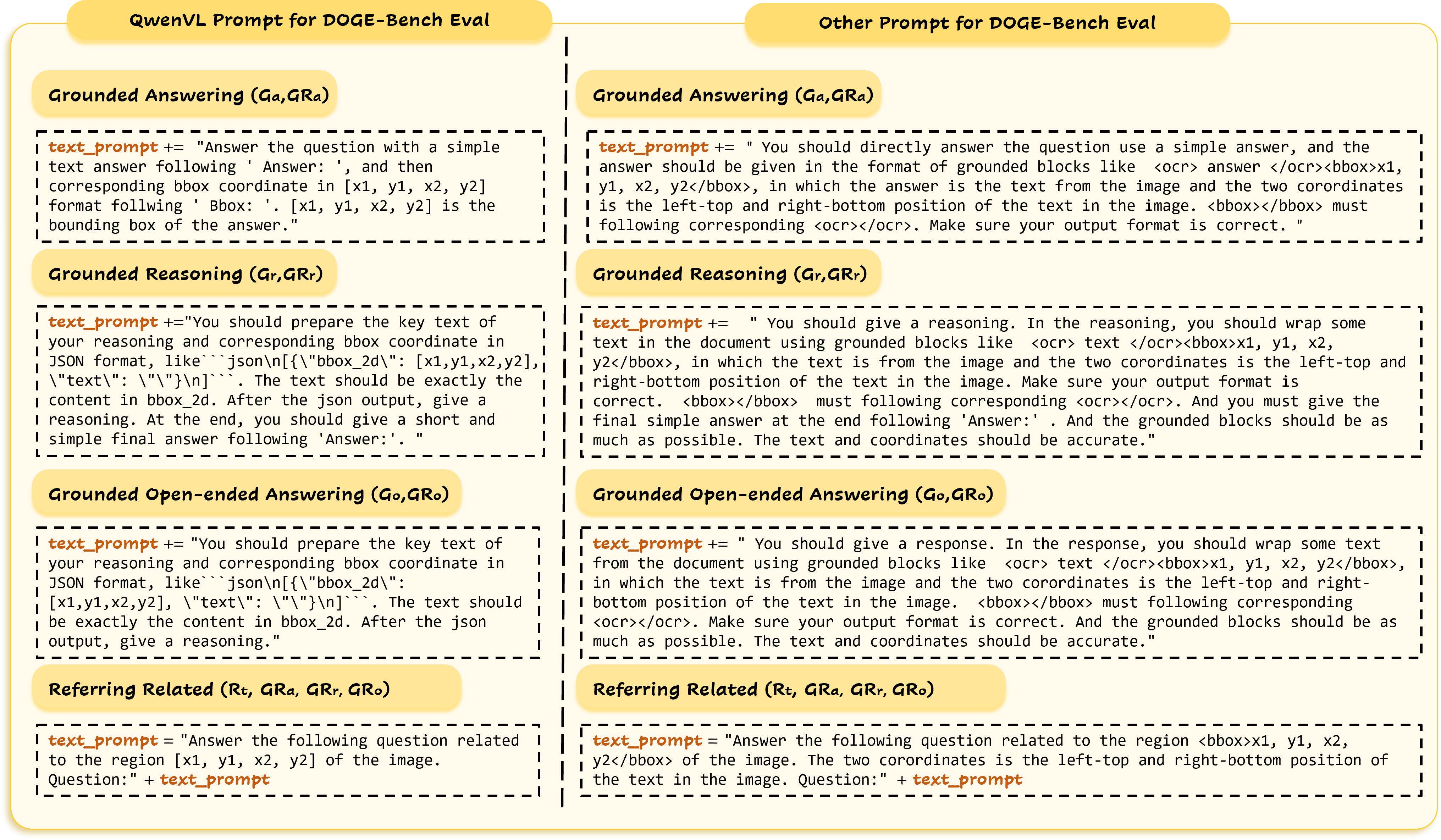}
   \caption{Prompts using in evaluation MLLMs on DOGR-Bench.}
   \label{fig:prompteval}

\end{figure*}

Since models other than InternVL2.5 and QwenVL2.5 do not have a clearly defined format for grounding or referring content, we directly interpret the input using the prompt shown in \cref{fig:prompteval}(Right) and restrict the output format for evaluation. For InternVL2.5, the use of \texttt{<ref></ref>} is required to trigger grounding capabilities. Therefore, when outputs require answers and corresponding coordinates, we additionally include \texttt{<ref></ref>} to obtain a more standardized output for evaluation. For QwenVL2.5, since the model outputs pixel coordinates of the input image rather than the coordinates normalized according to the respective width and height, we first convert the coordinates in our input to facilitate the model's accurate understanding of location information. We also restrict the output content to closely mimic the existing JSON format for evaluation purposes. For example, in reasoning tasks, the model first outputs a JSON list, then generates an explanation, and finally provides the answer. After obtaining the output, we normalize the output coordinates based on the input size for evaluation. The prompt we use for Qwen2.5 is shown in \cref{fig:prompteval}(Left).

\subsection{Limited Grounding Ability}
\begin{figure*}[]
  \centering
  \includegraphics[width=1\linewidth]{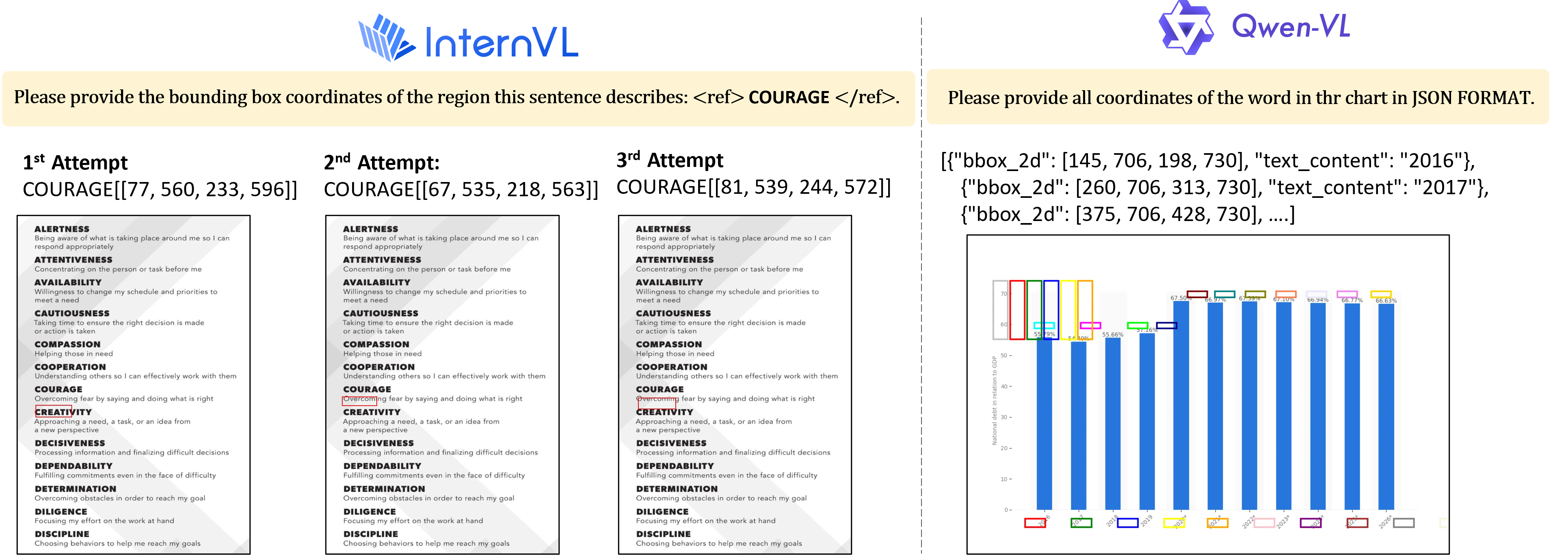}
   \caption{QwenVL2.5-72B and InternVL2.5-72B are all struggle with document grounding.}
   \label{fig:mllmfail}

\end{figure*}

Although both InternVL2.5\cite{chen2024expanding} and QwenVL2.5\cite{Qwen2.5-VL} can perform general image grounding, such as providing bounding boxes for a target based on a description, as shown in \cref{fig:mllmfail}, when handling document grounding, we use a grounding-triggering prompt to obtain the output coordinates corresponding to the given content. InternVL2.5-72B consistently outputs inaccurate and unstable bounding boxes, while QwenVL2.5 also produces a large number of inaccurate bounding boxes. These simple examples illustrate the models' shortcomings in document grounding capabilities.
\subsection{IoU Threshold Sensitivity}
\begin{figure}[]
  \centering
  \includegraphics[width=1\linewidth]{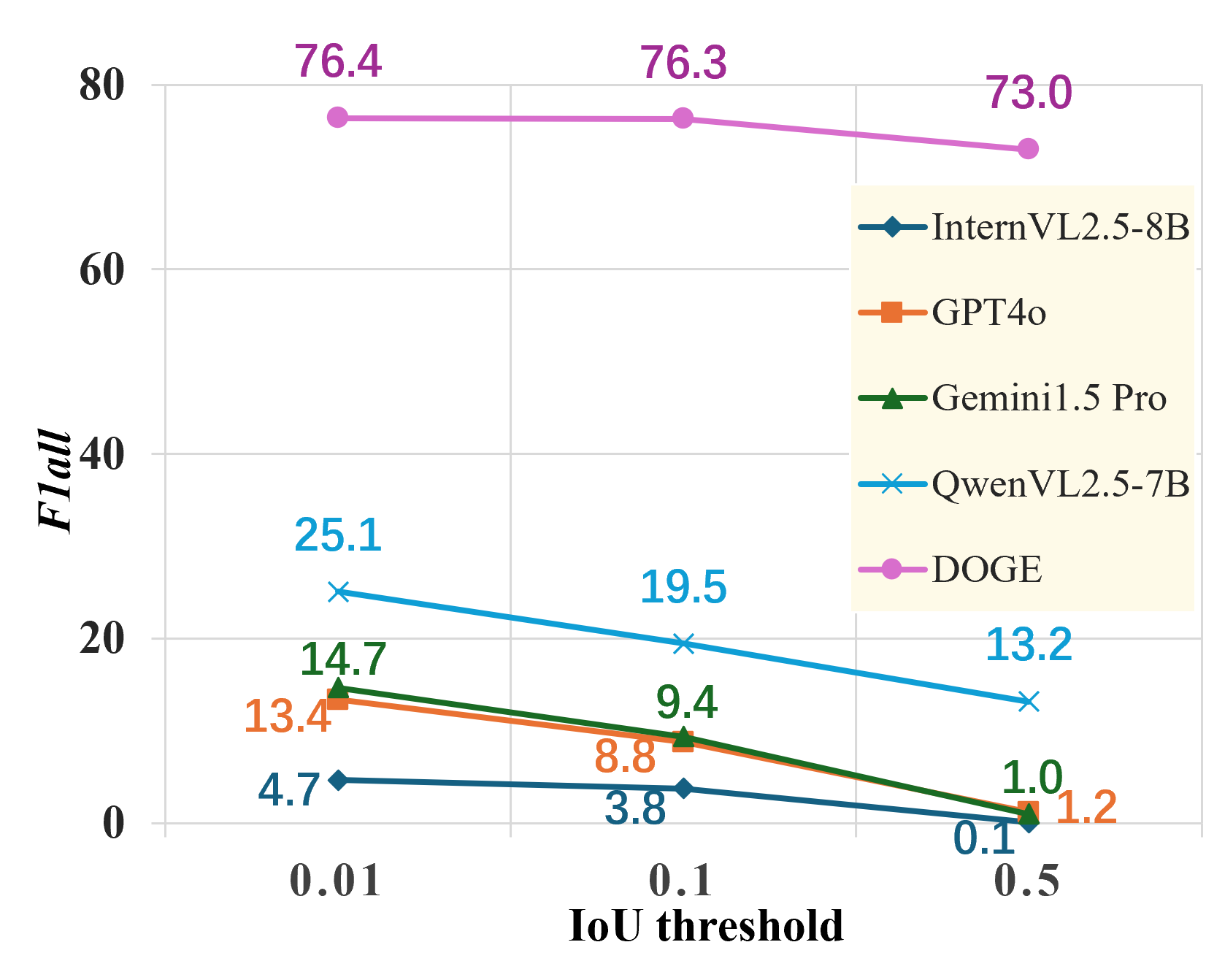}
   \caption{MLLM's $G_{a}$ ${F1_{all}}$ comparison on different IoU threshold. } 
   \label{fig:threshold}

\end{figure}

We also conduct a statistical analysis of the $G_a$ $F1_{all}$ scores of different MLLMs at various IoU thresholds. As shown in \cref{fig:threshold}, among the MLLMs analyzed, apart from DOGR, Qwen2.5VL-7B exhibits the best grounding capability. DOGR is not sensitive to changes in the IoU threshold, indicating that the limitation on DOGR's $G_a$ $F1_{all}$ is due to the model's reasoning capabilities. DOGR has already overcome the performance degradation caused by inaccurate grounding. In contrast, the insufficient basic grounding capabilities of other MLLMs make them very sensitive to changes in the IoU threshold, with their $G_a$ $F1_{all}$ scores dropping to nearly 0 when the IoU threshold is set to 0.5.

\section{More Experiments}

\noindent
{\bf Effectiveness of our pretraining data.} To better assess base performance of pretrained model using our multi-granular parsing data, we further introduce our text\&bbox localization test set \textbf{DOGR-Local15k} comprising four granularities (word, phrase, line, paragraph) across three categories of data: poster, chart, and general document, similar to DocLocal4k\cite{mplug1.5}. 
DOGR$^{-}$ is pre-trained without using our constructed multi-granular parsing data.
For the chart data, we claim that the chart type annotations in DocStruct4M for during DOGR$^{-}$'s pre-training are inconsistent with our grounding training objectives, which aims to match the text and the bbox. In the original data, the text corresponds to bounding boxes of bar/line charts. Therefore, these value is not referenceable, and we mark them in gray. Moreover, in the case where the granularity of data categories for charts is relatively singular, we directly reported the results for ``ALL''.

As shown in \cref{tab:pretrain_more}, after incorporating our multi-granular parsing data for pre-training, the model exhibits enhancements in text recognition and grounding tasks. Our improvements on DOGR-Local15k are significant, thereby laying a solid foundation for accurate document grounding interactions.

\noindent
{\bf The performance boost of data generated by GPT-4o.}
We utilize a weaker LLM, \textit{Gemini 1.5 flash}, to generate 25k chart grounding QA data (Gemini25k).
As shown in Tab.\ref{tab:chart_abl}, both Gemini25k and GPT-4o25k lead to improvement on ChartQA, which is mainly because constructed QA data enriches the original data. However, GPT-4o's stronger instruction-following capability results in a larger amount of grounded texts in the constructed QA, making it better than gemini-1.5-flash on ${G_r}$ task.  

\noindent
{\bf Importance of accurate bounding box annotation.}
To validate importance of accurate bounding box annotation, we conduct experiments on chart data. We perform random offset and scaling of bboxes in GPT4o data within 30\% to get 25k ${G_r}$ inaccurate training data GPT4o25k${^-}$ with errors for chart. We compare the performance gain of 25k rerendered data with accurate bboxes and 25k grounded data with inaccurate bboxes on ChartQA and ${G_r}$ of chart on DO-Bench. Both text are from GPT-4o. As shown in Tab.\ref{tab:chartdoge_abl}, adding GPT4o25k${^-}$, the performance on ChartQA is improved and the model possesses grounding capabilities. Adding accurate annotation data GPT4o25k${^-}$ leads to more precise bboxes, thus largely improving the grounding performance(F1 of ${G_r}$).

\begin{table}[h]
    \centering

        \caption{The performance boost of data generated by GPT-4o and weaker Gemini 1.5 flash.}

        \resizebox{0.99\linewidth}{!}{
        \begin{tabular}{c|c|cc}
            \toprule
            Training Data & ChartQA-Acc & Gr-Acc & Gr-F1\\ \midrule
            baseline(ChartQA) &80.92   &-& -      \\
+GPT4o25k&81.88 &67.0&    32.91  \\
+Gemini25k&81.98  &50.12& 1.10  \\ \bottomrule
        \end{tabular}
        \label{tab:chart_abl} 
        }

\end{table}

\begin{table}[h]

        \centering
        \caption{Accurate bbox is helpful for grounding.}
        \vspace{-11pt}
        
        \resizebox{0.96\linewidth}{!}{
        \begin{tabular}{c|c|c|c}
            \toprule
            Training Data & ChartQA-Acc & Gr-Acc & Gr-F1 \\ \midrule
            
            baseline(ChartQA) &80.92&-&-         \\
             +GPT4o25k${^-}$&81.42&66.1&20.2     \\  
            +GPT4o25k&81.88&67.0&32.91      \\
             
             \bottomrule
        \end{tabular}
        \label{tab:chartdoge_abl} 
        
        }
   
    \vspace{-13pt}
    
\end{table}
\end{CJK}
\end{document}